\documentclass[10pt,twocolumn,letterpaper]{article}

\usepackage{cvpr}
\usepackage{times}
\usepackage{epsfig}
\usepackage{lipsum}
\usepackage{graphicx}
\usepackage{amsmath}
\usepackage{amssymb}
\usepackage{balance}
\usepackage{subcaption}
\usepackage{multirow}

\usepackage[pagebackref=true,breaklinks=true,letterpaper=true,colorlinks,bookmarks=false]{hyperref}

\newcommand{\deva}[1]{\noindent \textcolor{red}{DEVA: #1}}

\cvprfinalcopy 

\def\cvprPaperID{314} 

\ifcvprfinal\fi

\begin{document}

\title{Finding Tiny Faces}

\author{Peiyun Hu, Deva Ramanan\\
  Robotics Institute\\
  Carnegie Mellon University\\
  {\tt\small \{peiyunh,deva\}@cs.cmu.edu}
}

\maketitle
\thispagestyle{empty}

\begin{abstract}
 Though tremendous strides have been made in object recognition, one of the remaining open challenges is detecting small objects. We explore three aspects of the problem in the context of finding small faces: the role of scale invariance,  image resolution, and contextual reasoning. While most recognition approaches aim to  be scale-invariant, the cues for recognizing a 3px tall face are fundamentally different than those for recognizing a 300px tall face. We take a different approach and train separate detectors for different scales. To maintain efficiency, detectors are trained in a {\em multi-task} fashion: they make use of features extracted from multiple layers of single (deep) feature hierarchy. While training detectors for large objects is straightforward, the crucial challenge remains training detectors for small objects. We show that context is crucial, and define templates that make use of {\em massively-large receptive fields} (where 99\% of the template extends beyond the object of interest). Finally, we explore the role of scale in pre-trained deep networks,  providing ways to extrapolate networks tuned for limited scales to rather extreme ranges. We demonstrate state-of-the-art results on massively-benchmarked face datasets (FDDB and WIDER FACE). In particular, when compared to prior art on WIDER FACE, our results {\bf reduce error by a factor of 2} (our models produce an AP of 82\% while prior art ranges from 29-64\%).
\end{abstract}

\section{Introduction}
\label{sec:intro}




\begin{figure*}[t!]
  \centering
  \includegraphics[width=\linewidth]{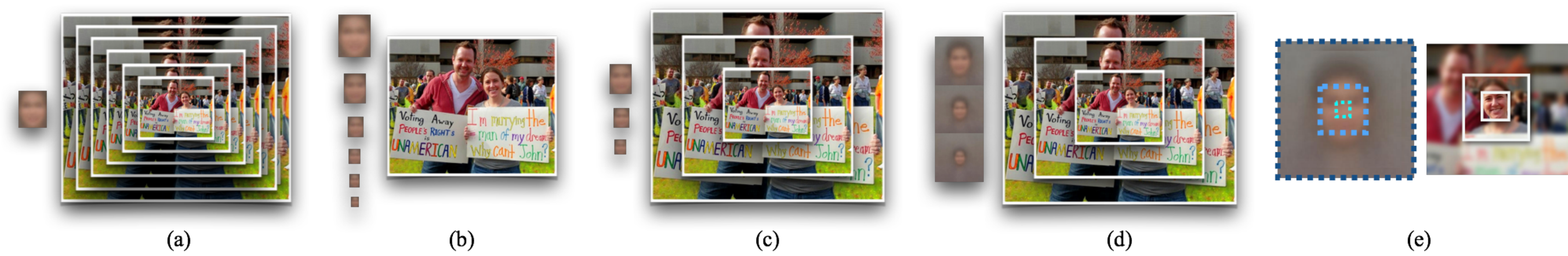}
  \caption{Different approaches for capturing scale-invariance. Traditional approaches build a single-scale template that is applied on a finely-discretized image pyramid (a). To exploit different cues available at different resolutions, one could build different detectors for different object scales (b). Such an approach may fail on extreme object scales that are rarely observed in training (or pre-training) data. We make use of a coarse image pyramid to capture extreme scale challenges in (c). Finally, to improve performance on small faces, we model additional context, which is efficiently implemented as a fixed-size receptive field across all scale-specific templates (d). We define templates over features extracted from multiple layers of a deep model, which is analogous to foveal descriptors (e). }
  \label{fig:overview}
\end{figure*}

Though tremendous strides have been made in object recognition, one of the remaining open challenges is detecting small objects. We explore three aspects of the problem in the context of face detection: the role of scale invariance, image resolution and contextual reasoning. Scale-invariance is a fundamental property of almost all current recognition and object detection systems. But from a practical perspective, scale-invariance cannot hold for sensors with finite resolution: the cues for recognizing a 300px tall face are undeniably different that those for recognizing a 3px tall face.

{\bf Multi-task modeling of scales:} Much recent work in object detection makes use of scale-normalized classifiers (e.g., scanning-window detectors run on an image pyramid~\cite{felzenszwalb2010object} or region-classifiers run on ``ROI''-pooled image features~\cite{girshick2014rich,ren2015faster}). When resizing regions to a canonical template size, we ask a simple question --{\it what should the
  size of the template be?} On one hand, we want a small template that
can detect small faces; on the other hand, we want a large template
that can exploit detailed features (of say, facial parts) to increase
accuracy. Instead of a ``one-size-fits-all'' approach, we train separate detectors tuned for different scales (and aspect ratios). Training a large collection of scale-specific detectors may suffer from lack of training data for individual scales and inefficiency from running a large number of detectors at test time. To address both concerns, we train and run scale-specific detectors in a {\em multi-task} fashion 
: they make use of features defined over multiple layers of single (deep) feature hierarchy. While such a strategy results in detectors of high accuracy for large objects, finding small things is still challenging.

{\bf How to generalize pre-trained networks?} We provide two remaining key insights to the problem of finding small objects. The first is an analysis of how best to extract scale-invariant features from pre-trained deep networks. We demonstrate that existing networks are tuned for objects of a characteristic size (encountered in pre-training datasets such as ImageNet).  To extend features fine-tuned from these networks to objects of novel sizes, we employ a simply strategy: resize images at test-time by interpolation and decimation. While many recognition systems are applied in a ``multi-resolution'' fashion by processing an image pyramid, we find that interpolating the lowest layer of the pyramid is particularly crucial for finding small objects~\cite{felzenszwalb2010object}. Hence our final approach  (Fig.~\ref{fig:overview}) is a delicate mixture of scale-specific detectors that are used in a scale-invariant fashion (by processing an image pyramid to capture large scale variations).

\begin{figure}[t]
  \centering
  \includegraphics[width=\linewidth]{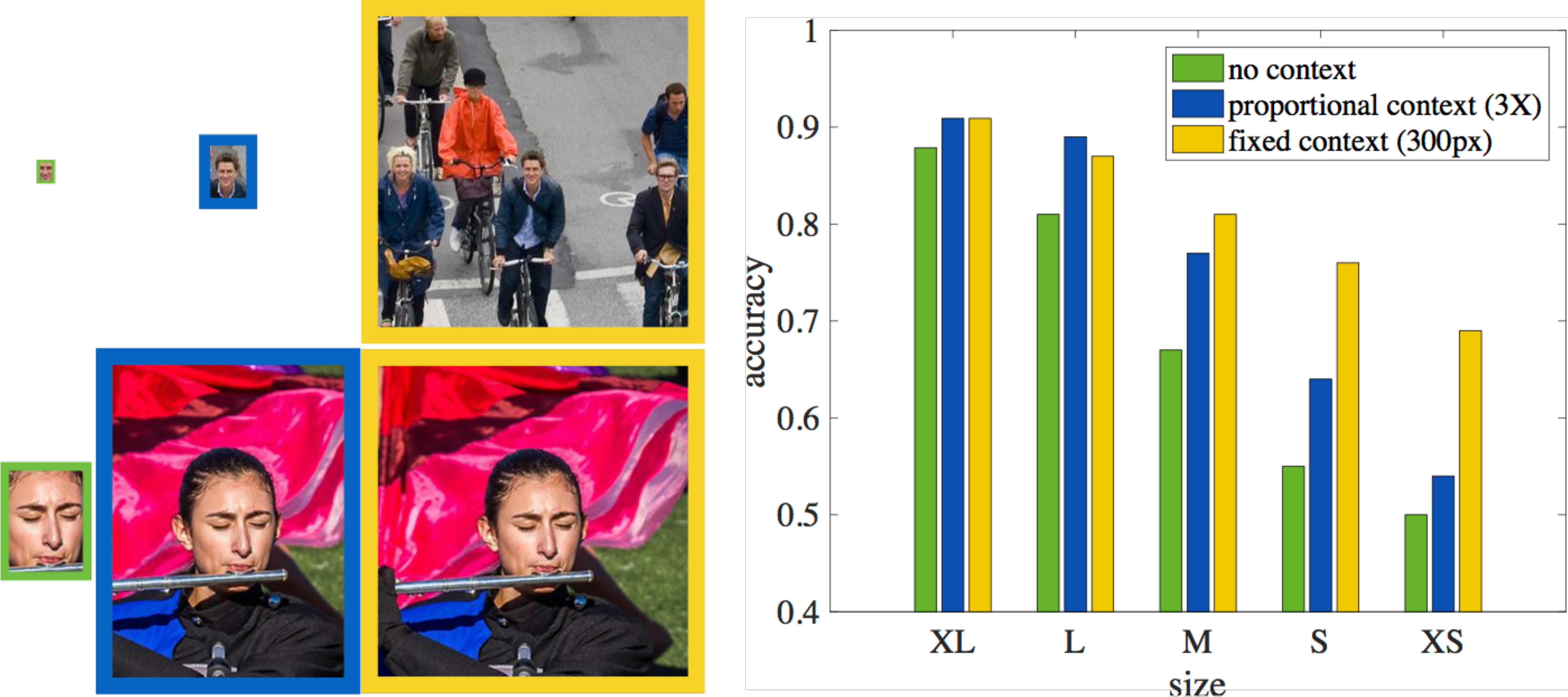}
  \caption{On the {\bf left}, we visualize a large and small face, both with and without context. One does not need context to recognize the large face, while the small face is dramatically unrecognizable without its context. We quantify this observation with a simple human experiment on the {\bf right}, where users classify true and false positive faces of our proposed detector.  Adding {\em proportional} context (by enlarging the window by 3X) provides a small improvement on large faces but is insufficient for small faces.  Adding a {\em fixed} contextual window of 300 pixels dramatically reduces error on small faces by 20\%.  This suggests that context should be modeled in a scale-{\em variant} manner. We operationalize this observation with foveal templates of massively-large receptive fields (around 300x300, the size of the yellow boxes).}
  \label{fig:context-example}  
\end{figure}

{\bf How best to encode context?} Finding small objects is fundamentally challenging because there is little signal on the object to exploit. Hence we argue that one must use image evidence beyond the object extent. This is often formulated as ``context''.   In Fig.~\ref{fig:context-example}, we present a simple human experiment where users attempt to classify true and false positive faces (as given by 
our detector). It is dramatically clear that humans need context to accurately classify small faces. Though this observation is quite intuitive and highly explored in computer vision~\cite{oliva2007role,torralba2003context}, it has been notoriously hard to quantifiably demonstrate the benefit of context in recognition~\cite{divvala2009empirical,galleguillos2010context,wolf2006critical}. One of the challenges appears to be how to effectively encode large image regions. We demonstrate that convolutional deep features extracted from multiple layers (also known as ``hypercolumn'' features \cite{hariharan2015hypercolumns,long2015fully}) are effective ``foveal'' descriptors that capture both high-resolution detail and coarse low-resolution cues across large receptive field (Fig.~\ref{fig:overview} (e)). 
We show that high-resolution components of our foveal descriptors (extracted from lower convolutional layers) are crucial for such accurate localization in Fig.~\ref{fig:context-case-foveal}.


{\bf Our contribution:} We provide an in-depth analysis of image resolution, object scale, and spatial context for the purposes of finding small faces. We demonstrate state-of-the-art results on massively-benchmarked face datasets (FDDB and WIDER FACE). In particular, when compared to prior art on WIDER FACE, our results {\bf reduce error by a factor of 2} (our models produce an AP of 82\% while prior art ranges from 29-64\%).


\section{Related work}
\label{sec:related}

{\bf Scale-invariance:}  The vast majority of recognition pipelines focus on scale-invariant representations, dating back to SIFT\cite{lowe2004distinctive}.  Current approaches to detection such as Faster RCNN~\cite{ren2015faster} subscribe to this philosophy as well,  extracting scale-invariant features through ROI pooling or an image pyramid~\cite{ren2015object}. We provide an in-depth exploration of scale-variant templates, which have been previously proposed for pedestrian detection\cite{park2010multiresolution}, sometimes in the context of improved speed~\cite{benenson2012pedestrian}. SSD~\cite{liu2015ssd} is a recent technique based on deep features that makes use of scale-variant templates. Our work differs in our exploration of context for tiny object detection.

{\bf Context:} Context is key to finding small instances as shown in multiple recognition tasks.
In object detection, \cite{bell2015inside} stacks spatial RNNs (IRNN\cite{le2015simple}) model context outside the region of interest and shows improvements on small object detection. In pedestrian detection, \cite{park2010multiresolution} uses ground plane estimation as contextual features and improves detection on small instances. In face detection, \cite{zhu2016cms} simultaneously pool ROI features around faces and bodies for scoring detections, which significantly improve overall performance. Our proposed work makes use of large local context (as opposed to a global contextual descriptor~\cite{bell2015inside,park2010multiresolution}) in a scale-variant way (as opposed to~\cite{zhu2016cms}). We show that context is mostly useful for finding low-resolution faces. 

{\bf Multi-scale representation:} Multi-scale representation has been proven useful for many recognition tasks. \cite{hariharan2015hypercolumns,long2015fully,BansalChen16} show that deep multi-scale descriptors (known as ``hypercolumns'') are useful for semantic segmentation. \cite{bell2015inside,liu2015ssd} demonstrate improvements for such models on object detection. \cite{zhu2016cms} pools multi-scale ROI features. Our model uses ``hypercolumn'' features, pointing out that fine-scale features are most useful for localizing small objects (Sec.~\ref{sec:context} and Fig.~\ref{fig:context-case-foveal}). 


{\bf RPN: } Our model superficially resembles a region-proposal network (RPN) trained for a specific object class instead of a general ``objectness'' proposal generator ~\cite{ren2015faster}. The important differences are that we use foveal descriptors (implemented through multi-scale features), we select a range of object sizes and aspects through cross-validation, and our models make use of an image pyramid to find extreme scales. In particular, our approach for finding small objects make use of scale-specific detectors tuned for interpolated images. Without these modifications, performance on small-faces dramatically drops by more than 10\% (Table~\ref{tab:scale-simplified}).



\section{Exploring context and resolution}
\label{sec:case}

In this section, we present an exploratory analysis of the issues at play that will inform our final model. To frame the discussion, we ask the following simple question: {\em what is the best way to find small faces of a fixed-size (25x20)?}. By explicitly factoring out scale-variation in terms of the desired output, we can explore the role of context and the canonical template size. Intuitively, context will be crucial for finding small faces. Canonical template size may seem like a strange dimension to explore - given that we want to find faces of size 25x20, why define a template of any size other than 25x20? 
Our analysis gives a surprising answer of when and why this should be done. To better understand the implications of our analysis, along the way we also ask the analogous question for a large object size: {\em what is the best way to find large faces of a fixed-size (250x200)?}.

{\bf Setup:} We explore different strategies for building scanning-window detectors for  fixed-size (e.g., 25x20) faces. We treat fixed-size object detection as a {\em binary heatmap prediction problem}, where the predicted heatmap at a pixel position $(x,y)$ specifies the confidence of a fixed-size detection centered at $(x,y)$. We train heatmap predictors using a fully convolutional network (FCN) ~\cite{long2015fully} defined over a state-of-the-art architecture ResNet~\cite{he2015deep}. We explore multi-scale features extracted from the last layer of each res-block, i.e. (res2cx, res3dx, res4fx, res5cx) in terms of ResNet-50. We will henceforth refer to these as (res2, res3, res4, res5) features. We discuss the remaining particulars of our training pipeline in Section~\ref{sec:exp}.

\begin{figure}
  \centering
  \includegraphics[width=.9\linewidth]{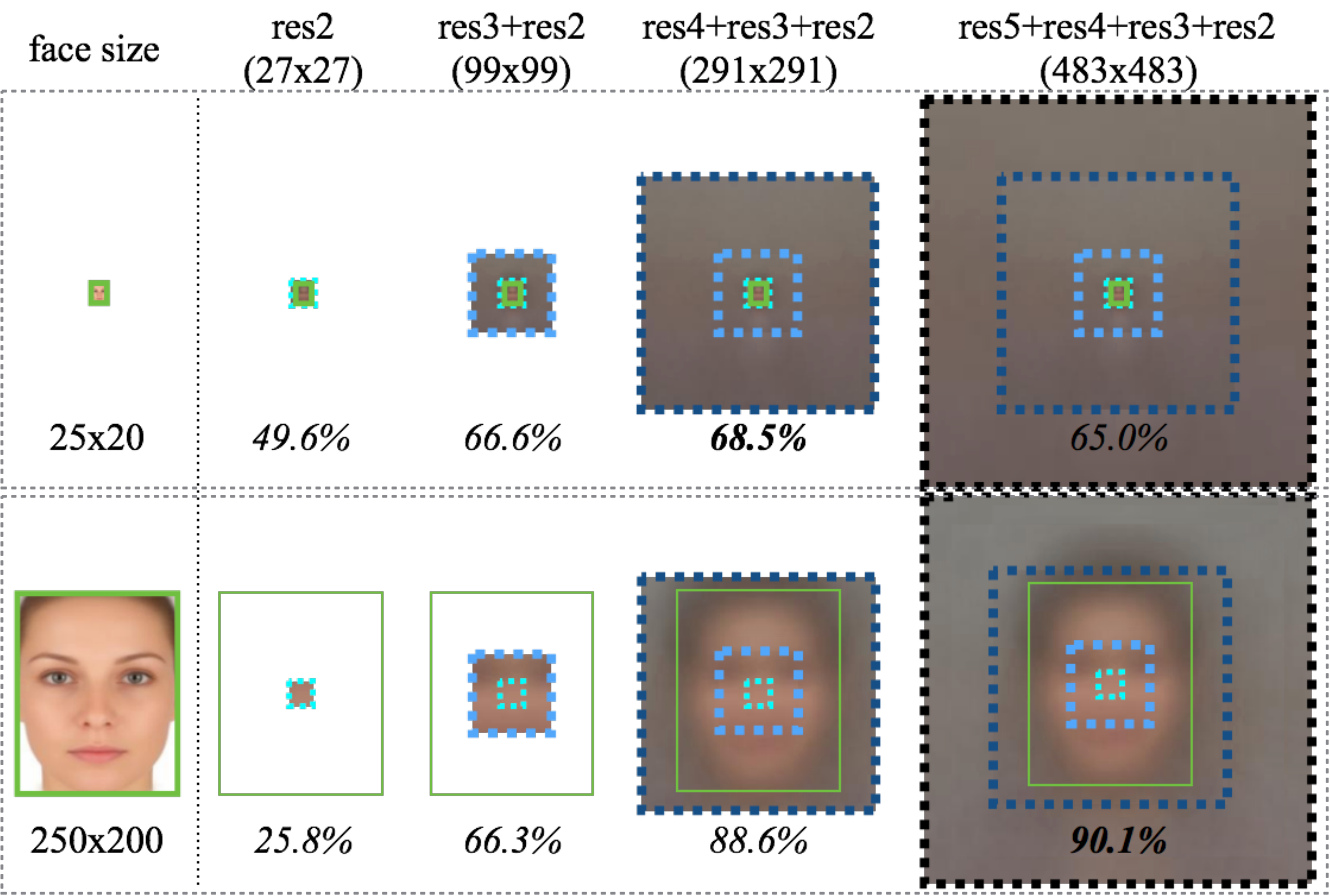}
  \caption{Modeling additional context helps, especially for finding small faces. The improvement from adding context to a tight-fitting template is greater for small faces (18.9\%) than for large faces (1.5\%). Interestingly smaller receptive fields do better for small faces, because the entire face is
visible. The green box represents the actual face size, while dotted boxes represent receptive fields associated with features from different layers (cyan = res2, light-blue = res3, dark-blue = res4, black = res5). Same colors are used in Figures~\ref{fig:context-case-foveal} and \ref{fig:scale-case}. }
  \label{fig:context-case}  
\end{figure}

\subsection{Context}
\label{sec:context}

\begin{figure}
  \centering
  \includegraphics[width=.9\linewidth]{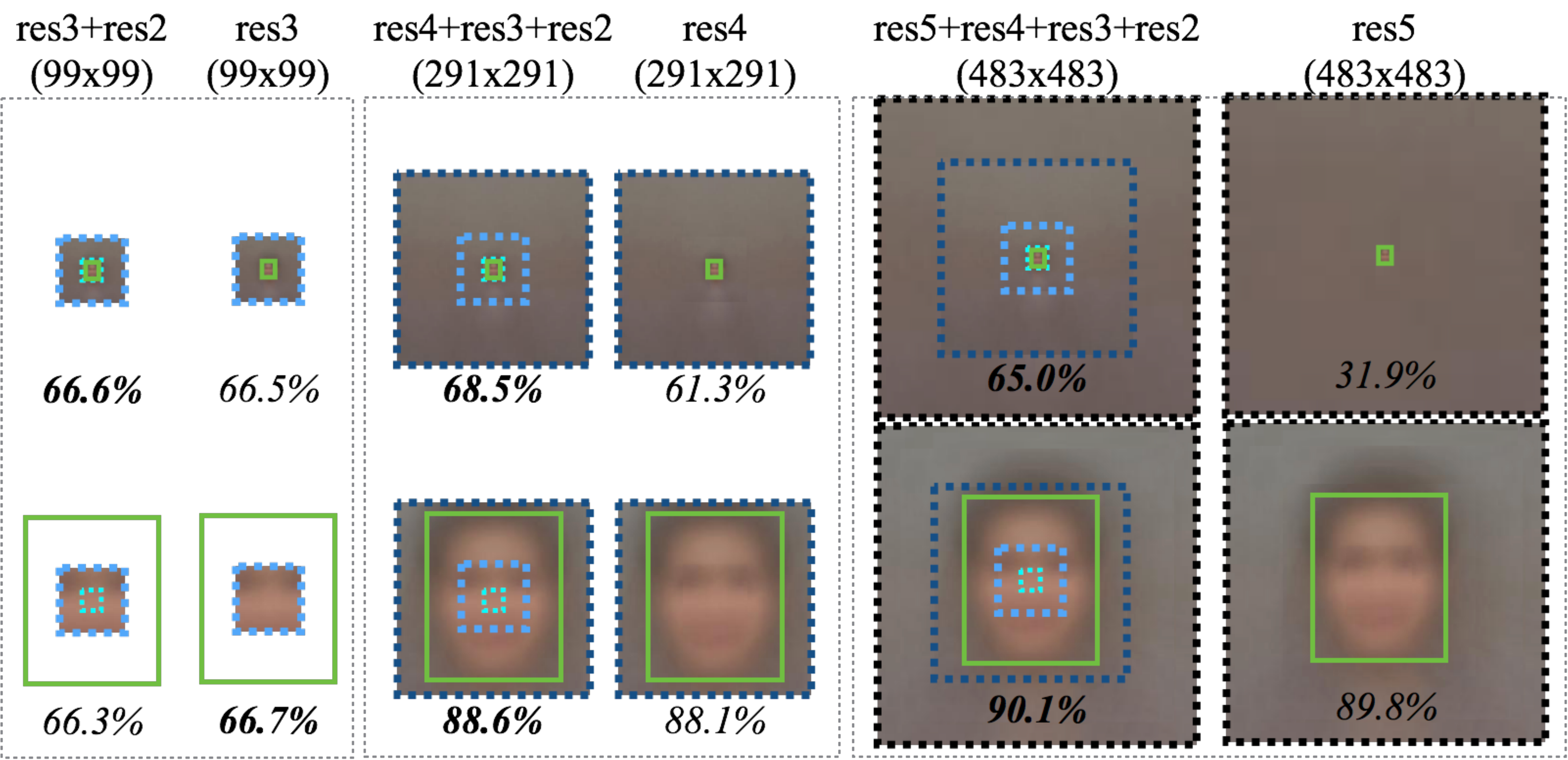}
  \caption{ Foveal descriptor is crucial for accurate detection on small objects. The small template ({\bf top}) performs 7\% worse with only res4 and 33\% worse with only res5. On the contrary, removing foveal structure does not hurt the large template ({\bf bottom}), suggesting high-resolution from lower layers is mostly useful for finding small objects!  }
  \label{fig:context-case-foveal}  
\end{figure}

Fig.~\ref{fig:context-case} presents an analysis of the effect of context, as given by the size of the receptive field (RF) used to make heatmap prediction. Recall that for fixed-size detection window, we can choose to make predictions using features with arbitrarily smaller or larger receptive fields compared to this window. Because convolutional features at higher layers tend to have larger receptive fields (e.g., res4 features span 291x291 pixels), smaller receptive fields necessitate the use of lower layer features. We see a number of general trends. Adding context almost always helps, though eventually additional context for tiny faces (beyond 300x300 pixels) hurts. We verified that this was due to over-fitting (by examining training and test performance). Interestingly, smaller receptive fields do better for small faces, because the entire face is visible - it is hard to find large faces if one looks for only the tip of the nose. More importantly, we analyze the impact of context by comparing performance of  a ``tight'' RF (restricted to the object extent) to the best-scoring ``loose''  RF with additional context. Accuracy for small faces improves by 18.9\%, while accuracy for large faces improves by 1.5\%, consistent with our human experiments (that suggest that context is most useful for small instances). Our results suggest that we can build multi-task templates for detectors of different sizes with identical receptive fields (of size 291x291), which is particularly simple to implement as a {\em multi-channel} heatmap prediction problem (where each scale-specific channel and pixel position has its own binary loss). In Fig.~\ref{fig:context-case-foveal}, we compare between descriptors with and without foveal structure, which shows that high-resolution components of our foveal descriptors are crucial for accurate detection on small instances. 
 

\subsection{Resolution}
\label{sec:resolution}

We now explore a rather strange question. What if we train a template whose size intentionally differs from the target object to be detected? In theory, one can use a ``medium''-size template (50x40) to find small faces (25x20) on a 2X upsampled (interpolated) test image. Fig.~\ref{fig:scale-case} actually shows the surprising result that this noticeably boosts performance, from 69\% to 75\%!  We ask the reverse question for large faces: can one find large faces (250x200) by running a template tuned for ``medium'' faces (125x100) on test images downsampled by 2X? Once again, we see a noticeable increase in performance, from 89\% to 94\%!

One explanation is that we have different amounts of training data for different object sizes, and we expect better performance for those sizes with more training data. A recurring observation in ``in-the-wild'' datasets such as WIDER FACE and COCO~\cite{lin2014microsoft} is that smaller objects greatly outnumber larger objects, in part because more small things can be labeled in a fixed-size image. We verify this for WIDER FACE in Fig.~\ref{fig:scale-overall} (gray curve). 
While imbalanced data may explain why detecting large faces is easier with medium templates (because there are more medium-sized faces for training), it does not explain the result for small faces. There exists {\em less} training examples of medium faces, yet performance is still much better using a medium-size template.

\begin{figure}
  \centering
  \includegraphics[width=.6\linewidth]{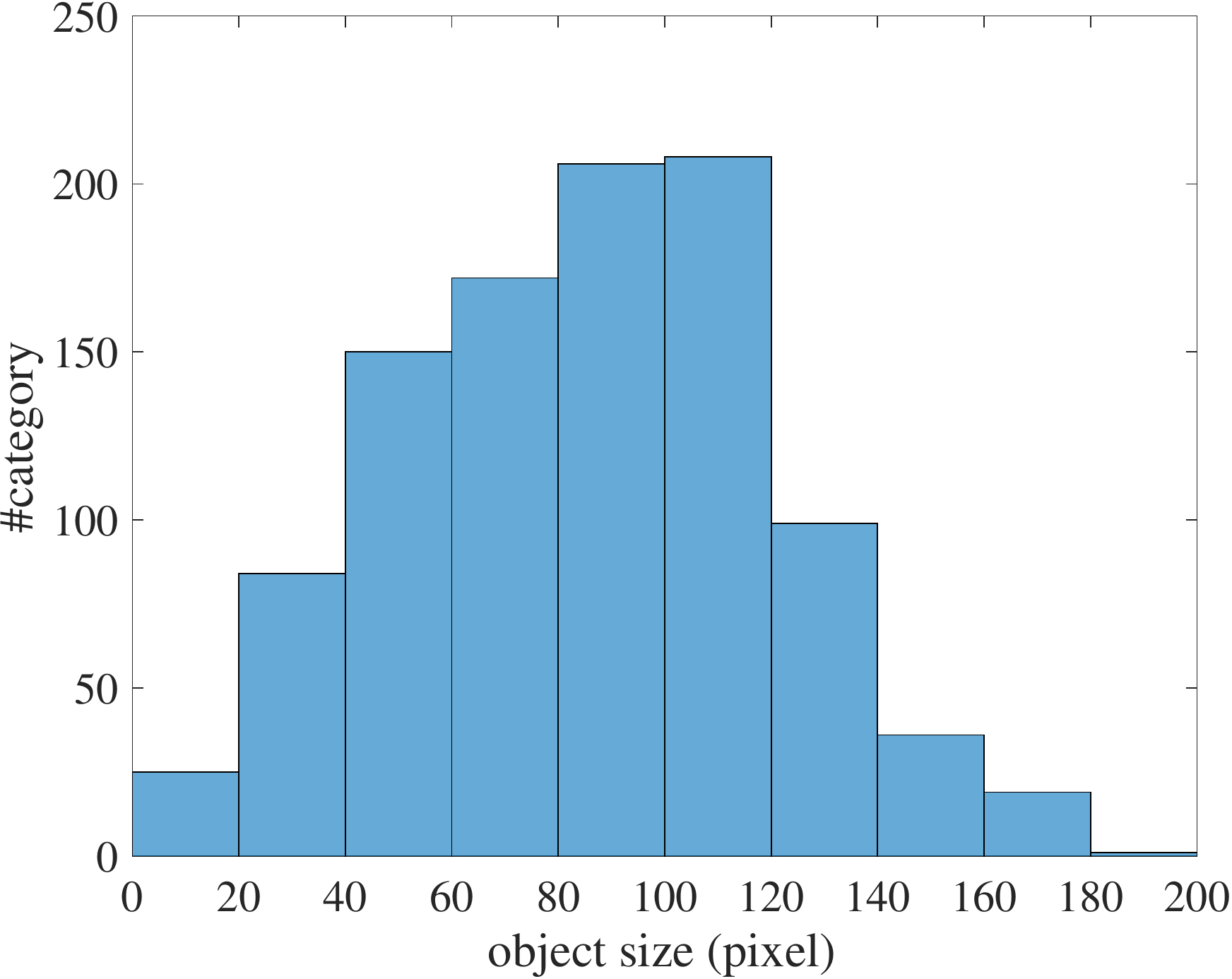}
  \caption{The distribution of average object scales in the ImageNet dataset (assuming images are normalized to 224x224). 
    More than 80\% categories have an average object size between 40 and 140 pixel. 
    We hypothesize that models pre-trained on ImageNet are optimized for objects in that range.  
  }
  \label{fig:imagenet}
\end{figure}

We find that the culprit lies in the distribution of object scales in the {\em pre-trained} dataset (ImageNet). Fig.~\ref{fig:imagenet} reveals that 80\% of the training examples in ImageNet contain objects of a ``medium'' size, between 40 to 140px. Specifically, we hypothesize that the pre-trained ImageNet model (used for fine-tuning our scale-specific detectors) is optimized for objects in that range, and that one should bias canonical-size template sizes to lie in that range when possible. We verify this hypothesis in the next section, where we describe a pipeline for building scale-specific detectors with varying canonical resolutions.


\section{Approach: scale-specific detection}
\label{sec:approach}


It is natural to ask a follow-up question: is there a general strategy for selecting template resolutions for particular object sizes? We demonstrate that one can make use of multi-task learning to ``brute-force'' train several templates at different resolution
, and greedily select the ones that do the best. As it turns out, there appears to be a general strategy consistent with our analysis in the previous section.

\begin{figure}[t]
  \centering
  \includegraphics[width=.9\linewidth]{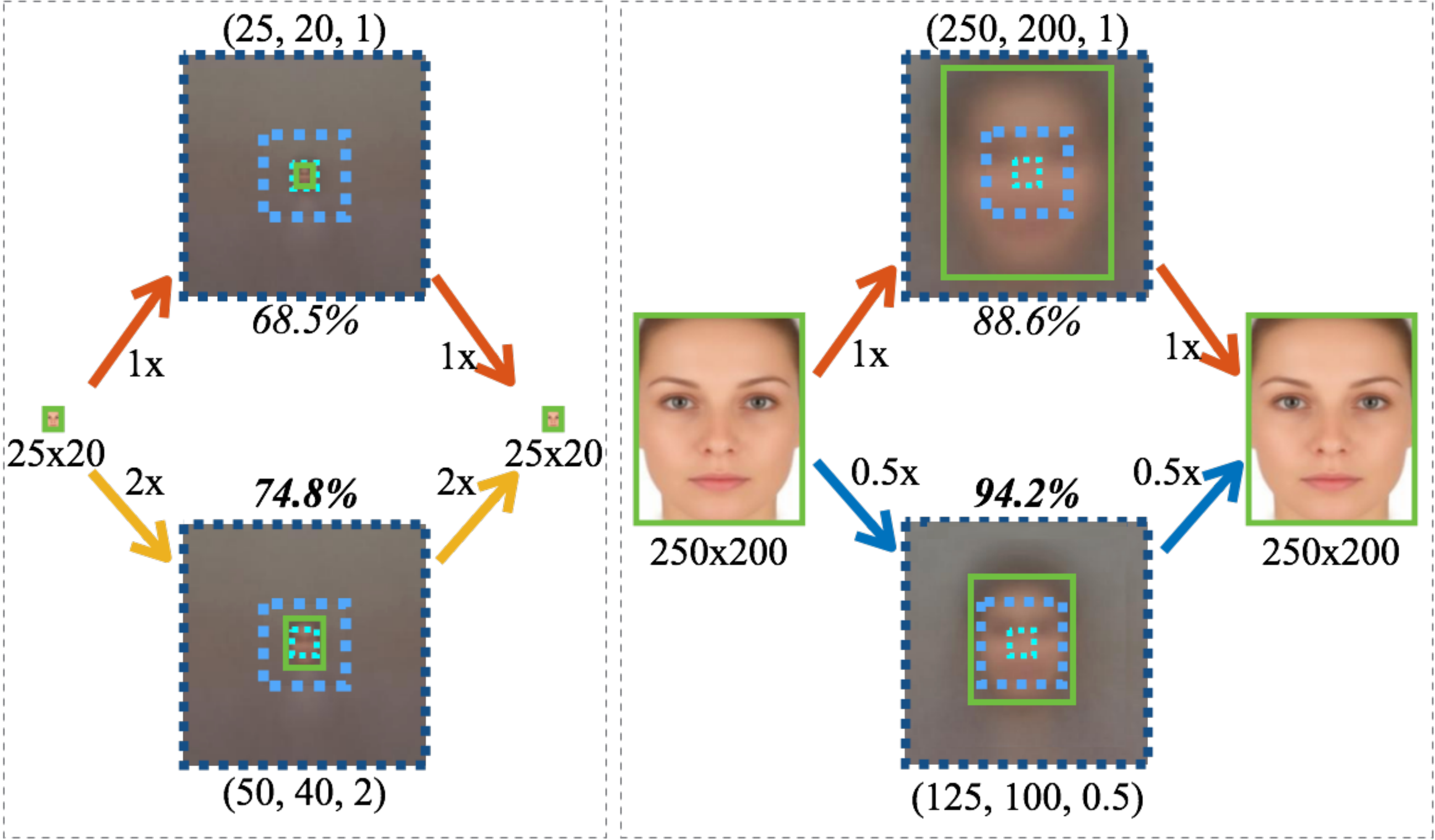}
  \caption{Building templates at original resolution is not optimal. For finding small (25x20) faces, building templates at 2x resolution improves overall accuracy by 6.3\%; while for finding large (250x200) faces, building templates at 0.5x resolution improves overall accuracy by 5.6\%. }
  \label{fig:scale-case}
\end{figure}

First, let us define some notation. We use $t(h, w, \sigma)$ to represent a template. Such a template is tuned to detect objects of size $(h/\sigma, w/\sigma)$ at resolution $\sigma$. For example, the right-hand-side Fig~\ref{fig:scale-case} uses both $t(250,200,1)$ (top) and $t(125,100,0.5)$ (bottom) to find 250x200 faces. 

Given a training dataset of images and bounding boxes, we can define a set of canonical bounding box shapes that roughly covers the bounding box shape space. In this paper, we define such canonical shapes by clustering, which is derived based on Jaccard distance $d$
(Eq.~\eqref{eq:pdist}):
\begin{equation}
  \label{eq:pdist}
  d(s_i,s_j) = 1 - \text{J}(s_i, s_j)
\end{equation}
where, $s_i=(h_i, w_i)$ and $s_j=(h_j, w_j)$ are a pair of bounding
box shapes and $J$ represents the standard Jaccard similarity (intersection over union overlap).

\begin{figure}[t]
  \centering
  \includegraphics[width=.9\linewidth]{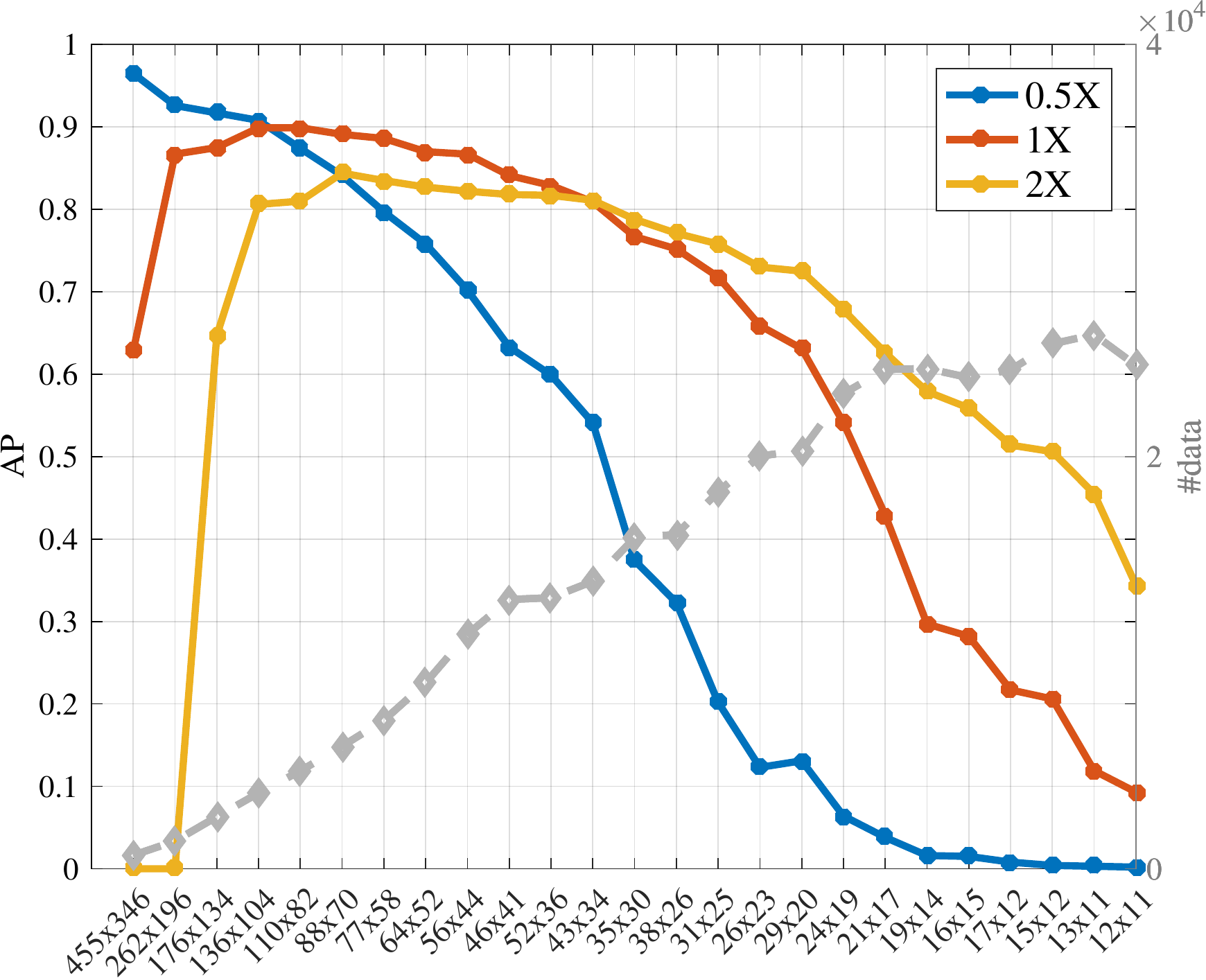}
  \caption{Template resolution analysis. X-axis represents target object sizes, derived by clustering. Left Y-axis shows AP at each target size (ignoring objects with more than 0.5 Jaccard distance). Natural regimes emerge in the figure: for finding large faces (more than 140px in height), build templates at 0.5 resolution; for finding smaller faces (less than 40px in height), build templates at 2X resolution. For sizes in between, build templates at 1X resolution. 
    Right Y-axis along with the gray curve shows the number of data within 0.5 Jaccard distance for each object size, suggesting that more small faces are annotated. }
  \label{fig:scale-overall}
\end{figure}

Now for each target object size $s_i=(h_i, w_i)$, we ask:
{\it what $\sigma_i$ will maximize performance of
  $t_i(\sigma_i h_i, \sigma_i w_i, \sigma_i)$?} 
To answer, we simply train separate multi-task models for each value of $\sigma \in \Sigma $ (some fixed set) and take the max for each object size. We plot the performance of each resolution-specific multi-task model as a colored curve in Fig.~\ref{fig:scale-overall}. With optimal $\sigma_i$ for each $(h_i, w_i)$, we retrain one multi-task model with ``hybrid'' resolutions (referred to as HR), which in practice follows the upper envelope of all the curves. Interestingly, there exist natural regimes for different strategies: to find large objects (greater than 140px in height),  use 2X smaller canonical resolution. To find small objects (less than 40px in height), use 2X larger canonical template resolution. Otherwise, use the same (1X) resolution. Our results closely follow the statistics of ImageNet (Fig.~\ref{fig:imagenet}), for which most objects fall into this range.

\begin{figure}[t]
  \centering
  \includegraphics[width=.6\linewidth]{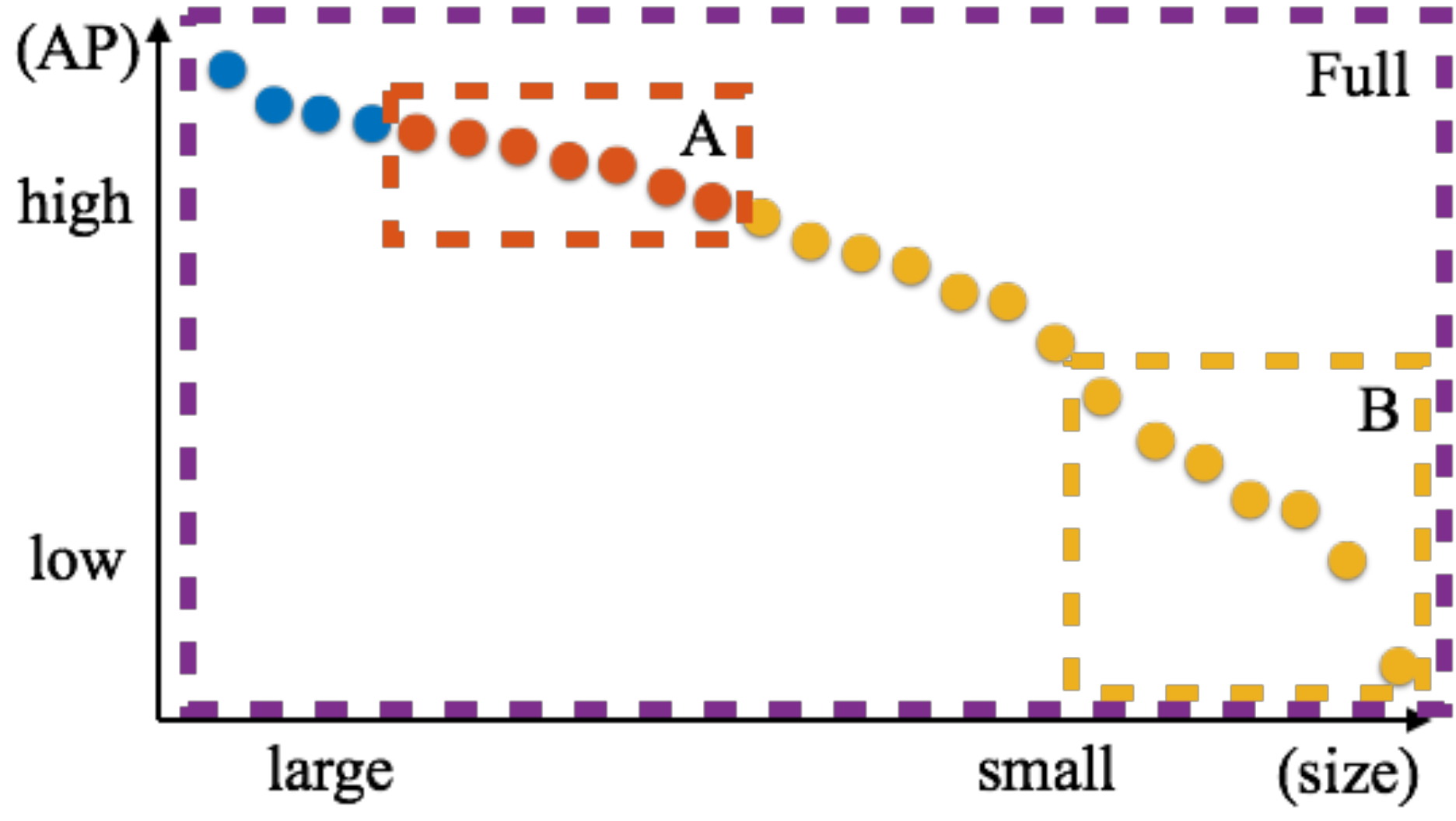}
  \caption{Pruning away redundant templates. Suppose we test templates built at 1X resolution (A) on a coarse image pyramid (including 2X interpolation). They will cover a larger range of scale except extremely small sizes, which are best detected using templates built at 2X, as shown in Fig.~\ref{fig:scale-overall}. Therefore, our final model can be reduced to two small sets of scale-specific templates: (A) tuned for 40-140px tall faces and are run on a coarse image pyramid (including 2X interpolation) and (B) tuned for faces shorter than 20px and are only run in 2X interpolated images. }
  \label{fig:scale-simplified}
\end{figure}

{\bf Pruning:} The hybrid-resolution multitask model in the previous section is somewhat redundant. 
For example, template $(62,50,2)$, the optimal template for finding 31x25 faces, is redundant given the existence of template $(64,50,1)$, the optimal template for finding 64x50 faces.
Can we prune away such redundancies? Yes! We refer the reader to the caption in Fig.~\ref{fig:scale-simplified} for an intuitive description. As Table~\ref{tab:scale-simplified} shows, pruning away redundant templates led to some small improvement. Essentially, our model can be reduced to a small set of scale-specific templates (tuned for 40-140px tall faces) that can be run on a coarse image pyramid (including 2X interpolation), combined with a set of scale-specific templates designed for finding small faces (less than 20px in height) in 2X interpolated images.

\begin{table}
  \centering
  \resizebox{.9\linewidth}{!}{
    \begin{tabular}{l|ccc}
      Method & Easy   & Medium & Hard   \\
      \hline \hline
      RPN & 0.896 & 0.847 & 0.716 \\
      \hline
      HR-ResNet101 (Full) & 0.919 & 0.908 & 0.823 \\
      HR-ResNet101 (A+B) & \textbf{0.925} & \textbf{0.914} & \textbf{0.831} \\
      \hline
    \end{tabular}
  }
  \caption{Pruning away redundant templates does not hurt performance (validation). As a reference, we also included the performance of a vanilla RPN as mentioned in Sec.~\ref{sec:related}. Please refer to Fig.~\ref{fig:scale-simplified} for visualization of (Full) and (A+B).} 
  \label{tab:scale-simplified}
\end{table}

\subsection{Architecture}
\label{sec:net}

\begin{figure*}[t]
  \centering
  \includegraphics[width=.8\linewidth]{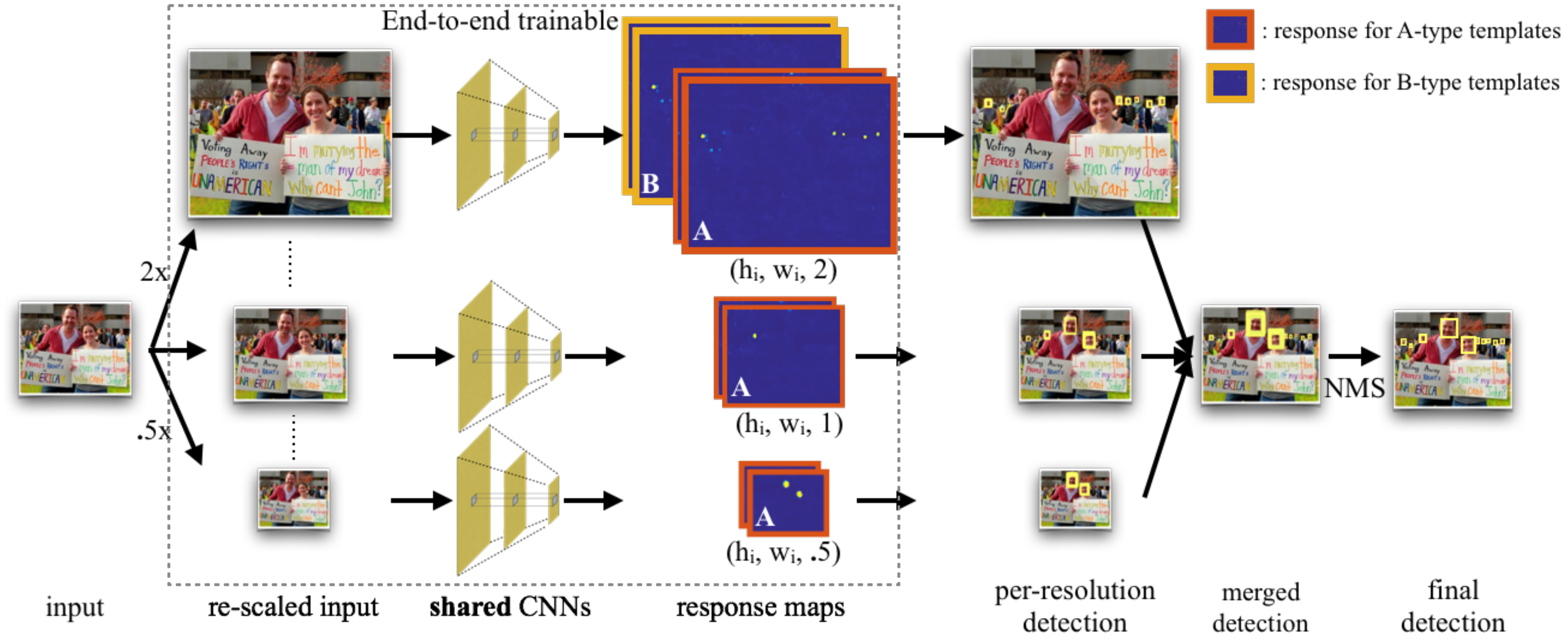}
  \caption{Overview of our detection pipeline. Starting with an input image, we first create a coarse image pyramid (including 2X interpolation). We then feed the scaled input into a CNN to predict template responses (for both detection and regression) at every resolution. In the end, we apply non-maximum suppression (NMS) at the original resolution to get the final detection results. The dotted box represents the end-to-end trainable part. We run A-type templates (tuned for 40-140px tall faces) on the coarse image pyramid (including 2X interpolation), while only run B-type (tuned for less than 20px tall faces) templates on only 2X interpolated images (Fig.~\ref{fig:scale-simplified})}
  \label{fig:arch}
\end{figure*}

We visualize our proposed architecture in Fig.~\ref{fig:arch}. We train binary multi-channel heatmap predictors to report object confidences for a range of face sizes (40-140px in height). We then find larger and smaller faces with a coarse image pyramid, which importantly includes a 2X upsampling stage with special-purpose heatmaps that are predicted only for this resolution (e.g., designed for tiny faces shorter than 20 pixels). For the shared CNNs, we experimented with ResNet101, ResNet50, and VGG16. Though ResNet101 performs the best, we included performance of all models in Table~\ref{tab:arch}. We see that 
{\em all} models achieve substantial improvement on ``hard'' set over prior art, including CMS-RCNN\cite{zhu2016cms} , which also models context, but in a proportional manner (Fig. \ref{fig:context-example}). 

\begin{table}
  \centering
  \resizebox{\linewidth}{!}{
    \begin{tabular}{l|ccc}
      Method & Easy   & Medium & Hard   \\
      \hline \hline
      ACF\cite{yang2014aggregate} & 0.659 & 0.541 & 0.273 \\
      Two-stage CNN\cite{yang2016wider} & 0.681 & 0.618 & 0.323 \\
      Multiscale Cascade CNN\cite{yang2015facial} & 0.691 & 0.634 & 0.345 \\
      Faceness\cite{yang2015facial} & 0.713 & 0.664 & 0.424 \\
      Multitask Cascade CNN\cite{zhang2016joint} & 0.848 & 0.825 & 0.598 \\
      CMS-RCNN\cite{zhu2016cms} & 0.899 & 0.874 & 0.624 \\
      \hline 
      HR-VGG16 & 0.862 & 0.844 & 0.749 \\
      HR-ResNet50 & 0.907 & 0.890 & 0.802 \\
      HR-ResNet101 & {\bf 0.919} & {\bf 0.908} & {\bf 0.823} \\
    \end{tabular}
  }
  \caption{Validation performance of our models with different {\em architectures}. ResNet101 performs slightly better than ResNet50 and much better than VGG16. Importantly, our VGG16-based model already outperforms prior art by a large margin on ``hard'' set. }
  \label{tab:arch}
\end{table}


{\bf Details:} Given training images with ground-truth annotations of objects and templates, 
we define positive locations to be those where IOU overlap exceeds 70\%, and negative locations to be those where the overlap is below 30\% (all other locations are ignored by zero-ing out the gradient 
). Note that this implies that each large object instance generates many more positive training examples than small instances. Since this results in a highly imbalanced binary classification training set, we make use of balanced sampling~\cite{girshick2014rich} and hard-example mining~\cite{shrivastava2016training} to ameliorate such effects. We find performance increased with a post-processing linear regressor that fine-tuned reported bounding-box locations. To ensure that we train on data similar to test conditions, we randomly resize training data to the range of $\Sigma$ resolution that we consider at test-time (0.5x,1x,2x) and learn from a fixed-size random crop of 500x500 regions per image (to take advantage of batch processing). We fine-tune pre-trained ImageNet models on the WIDER FACE training set with a fixed learning rate of $10^{-4}$, and evaluate performance on the WIDER FACE validation set (for diagnostics) and held-out testset. To generate final detections, we apply standard NMS to the detected heatmap with an overlap threshold of 30\%. We discuss more training details of our procedure in the Appendix~\ref{sec:experimental-details}. Both our code and models are available online at {\small \url{https://www.cs.cmu.edu/~peiyunh/tiny}}. 

\section{Experiments}
\label{sec:exp}

\begin{figure*}
  \centering
  \begin{minipage}{.32\textwidth}
    \centering
    \includegraphics[width=.89\linewidth]{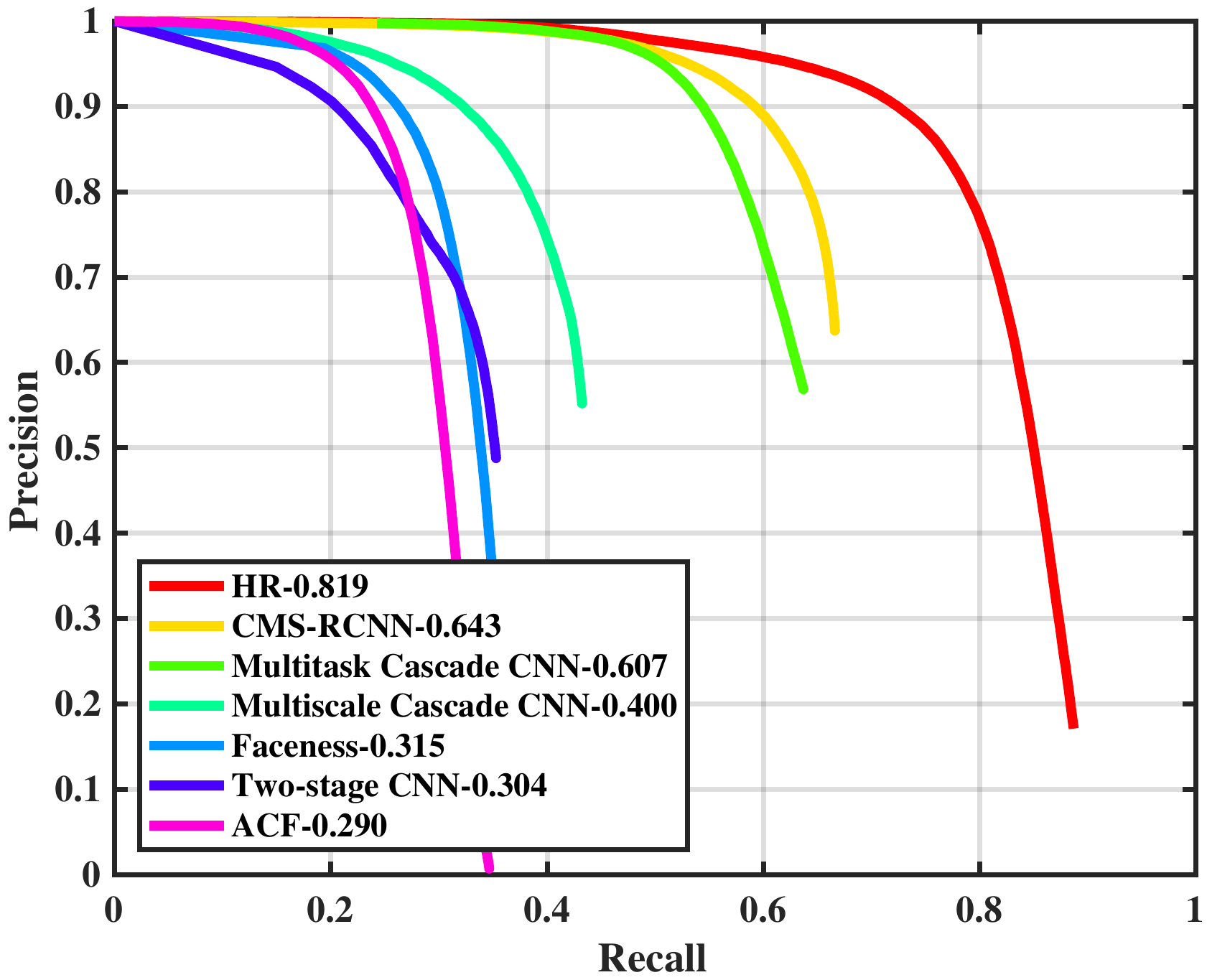}
    \captionof{figure}{Precision recall curves on WIDER FACE ``hard'' testset. Compared to prior art, our approach (HR) reduces error by 2X.}
    \label{fig:widerface}  
  \end{minipage} \hspace{1em}
  \begin{minipage}{.65\textwidth}
    \centering
    \includegraphics[width=\linewidth]{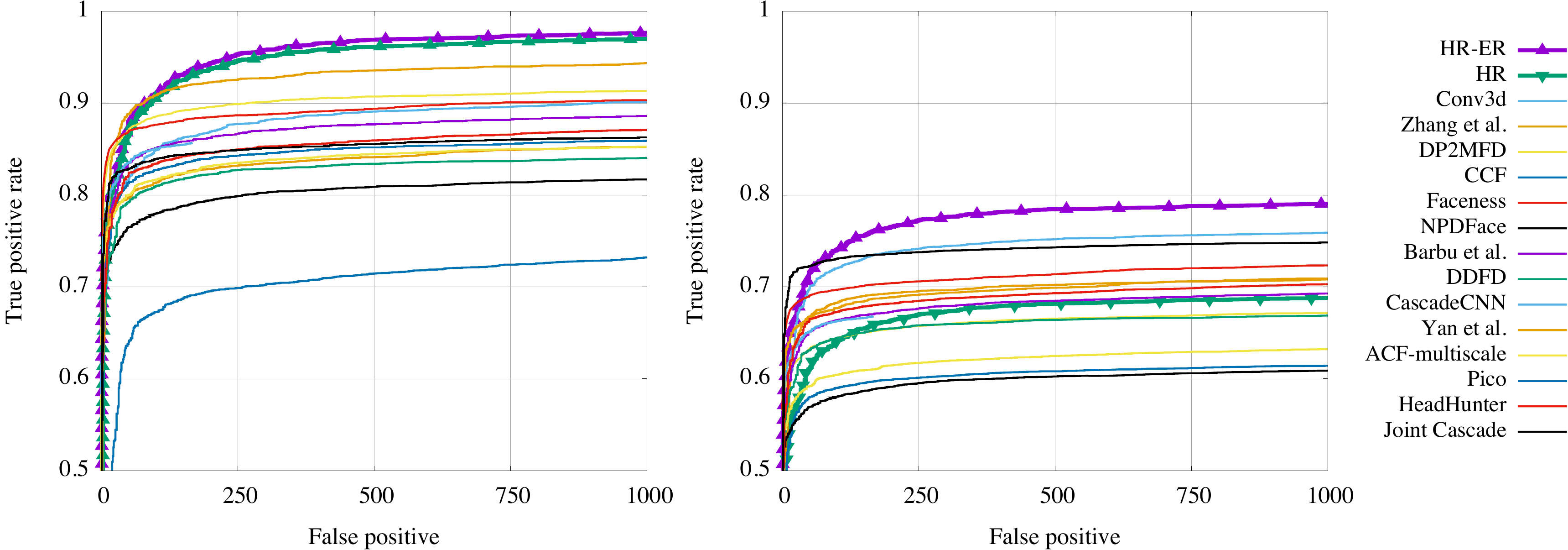}
    \captionof{figure}{ROC curves on FDDB-test. Our pre-trained detector (HR) produces state-of-the-art discrete detections ({\bf left}). By learning a post-hoc regressor that converts bounding boxes to ellipses, our approach (HR-ER) produces state-of-the-art continuous overlaps as well ({\bf right}). We compare to only published results.}
    \label{fig:fddb}  
  \end{minipage}
\end{figure*}

\begin{figure*}
  \centering
  \includegraphics[width=.995\linewidth]{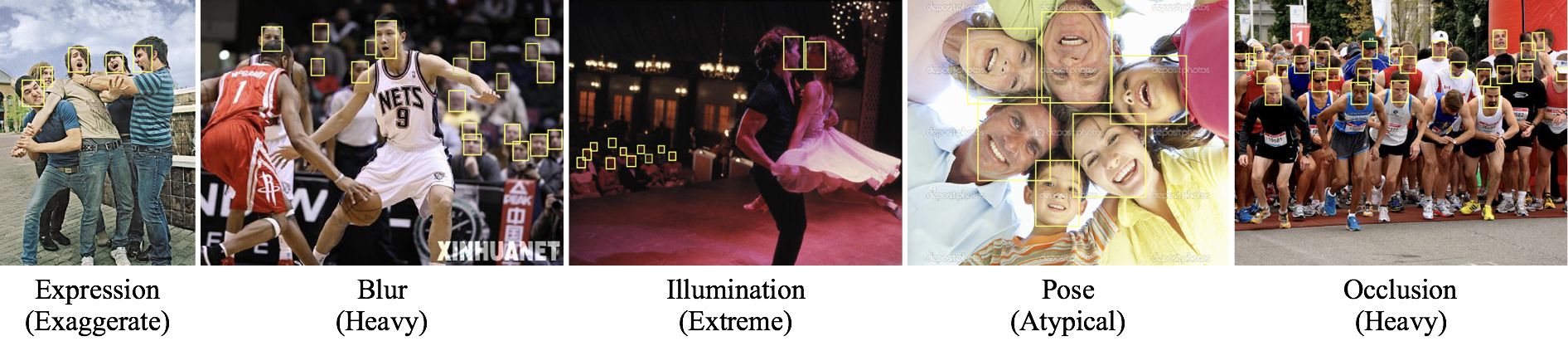}
  \includegraphics[width=.995\linewidth]{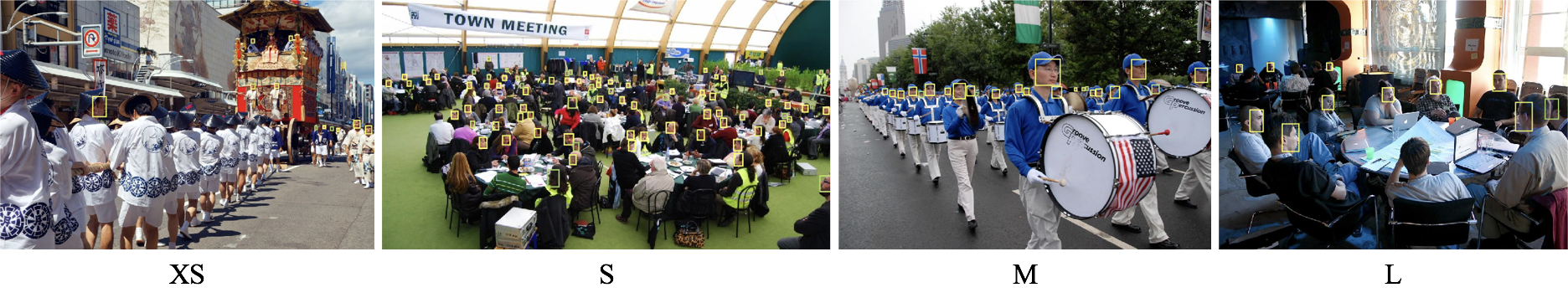}
  \caption{Qualitative results on WIDER FACE. We visualize one example for each attribute and scale. Our proposed detector is able to detect faces at a continuous range of scales, while being robust to challenges such as expression, blur, illumination etc. Please zoom in to look for some very small detections. }
  \label{fig:wider-example}
\end{figure*}

{\bf WIDER FACE:} We train a model with 25 templates on WIDER FACE's training set and report the performance of our best model {\em HR-ResNet101 (A+B)} on the held-out test set. 
As Fig.~\ref{fig:widerface} shows, our hybrid-resolution model (HR) achieves state-of-the-art performance on all difficulty levels, but most importantly, reduces error on the ``hard'' set by 2X. 
Note that ``hard'' set includes {\em all} faces taller than 10px, hence more accurately represents performance on the full testset. We visualize our performance under some challenging scenarios in Fig.~\ref{fig:wider-example}. Please refer to the benchmark website for full evaluation and our Appendix~\ref{sec:error-analysis} for more quantitative diagnosis~\cite{hoiem2012diagnosing}.  

\begin{figure*}
  \centering
  \includegraphics[width=\linewidth]{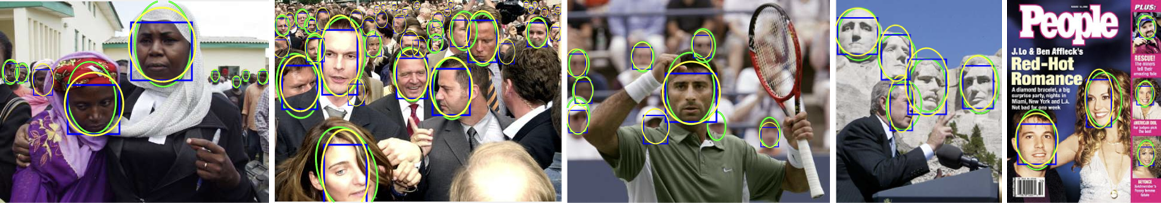}
  \caption{Qualitative results on FDDB. Green ellipses are ground truth, blue bounding boxes are detection results, and yellow ellipses are regressed ellipses. Our proposed detector is robust to heavy occlusion, heavy blur, large appearance and scale variance. Interestingly, many faces under such challenges are not even annotated (second example).}
  \label{fig:fddb-example}
\end{figure*}

{\bf FDDB:} We test our WIDER FACE-trained model on FDDB. Our out-of-the-box detector (HR) outperforms all published results on the discrete score, which uses a standard 50\% intersection-over-union threshold to define correctness. Because FDDB uses bounding ellipses while WIDER FACE using bounding boxes, we train a post-hoc linear regressor to transform bounding box predictions to ellipses. With the post-hoc regressor, our detector achieves state-of-the-art performance on the continuous score (measuring average bounding-box overlap) as well. Our regressor is trained with 10-fold cross validation. Fig.~\ref{fig:fddb} plots the performance of our detector both with and without the elliptical regressor (ER). Qualitative results are shown in Fig.~\ref{fig:fddb-example}. Please refer to our Appendix~\ref{sec:experimental-details} for a formulation of our elliptical regressor.

{\bf Run-time:} Our run-time is dominated by running a ``fully-convolutional'' network across a 2X-upsampled image. Our Resnet101-based detector runs at 1.4FPS on 1080p resolution and 3.1FPS on 720p resolution. Importantly, our run-time is {\em independent} of the number of faces in an image. This is in contrast to proposal-based detectors such as Faster R-CNN~\cite{ren2015faster}, which scale {\em linearly} with the number of proposals.

{\bf Conclusion:} We propose a simple yet effective framework for finding small objects, demonstrating that both large context and scale-variant representations are crucial.  We specifically show that massively-large receptive fields can be effectively encoded as a foveal descriptor that captures both coarse context (necessary for detecting small objects) and high-resolution image features (helpful for localizing small objects).
We also explore the encoding of scale in existing pre-trained deep networks, suggesting a simple way to extrapolate networks tuned for limited scales to more extreme scenarios in a scale-variant fashion. 
Finally, we use our detailed analysis of scale, resolution, and context to develop a state-of-the-art face detector that significantly outperforms prior work on standard benchmarks.

{\bf Acknowledgments:} This research is based upon work supported in part by NSF Grant 1618903, the Intel Science and Technology Center for Visual Cloud Systems (ISTC-VCS), Google, and the Office of the Director of National Intelligence (ODNI), Intelligence Advanced Research Projects Activity (IARPA), via IARPA R \& D Contract No. 2014-14071600012. The views and conclusions contained herein are those of the authors and should not be interpreted as necessarily representing the official policies or endorsements, either expressed or implied, of ODNI, IARPA, or the U.S. Government. The U.S. Government is authorized to reproduce and distribute reprints for Governmental purposes notwithstanding any copyright annotation thereon.

\appendix

\section{Error analysis}
\label{sec:error-analysis}

\begin{figure}
  \centering
  \includegraphics[width=.9\linewidth]{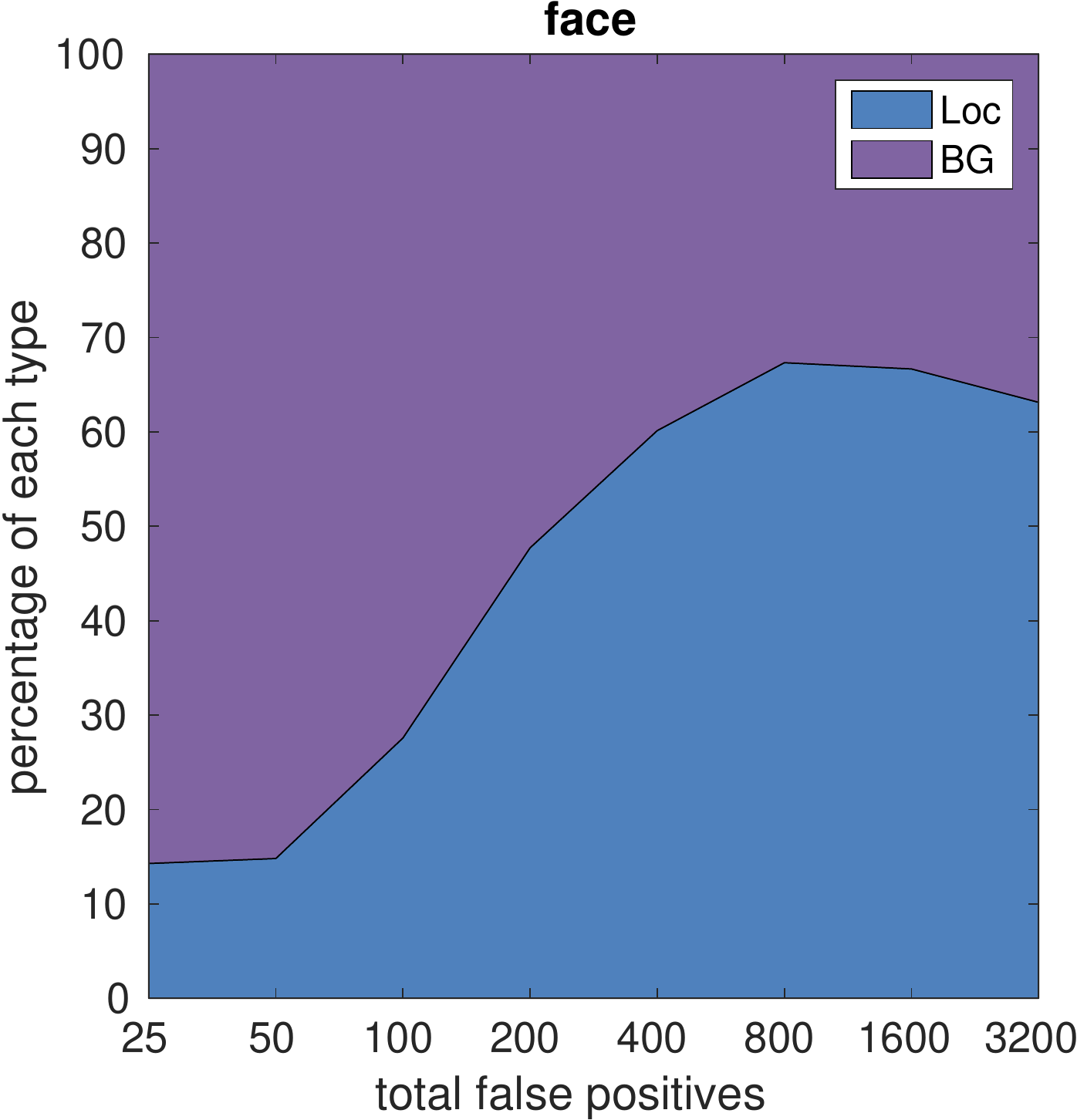}
  \caption{\label{fig:error} Distribution of error modes of false positives. Background confusion seems the dominating error mode among top-scoring detection, however, we found 15 out of 20 top-scoring false positives, as shown in Fig.~\ref{fig:topfp}, are in fact due to missed annotation. }
\end{figure}

\begin{figure}
  \centering
  \includegraphics[width=.9\linewidth]{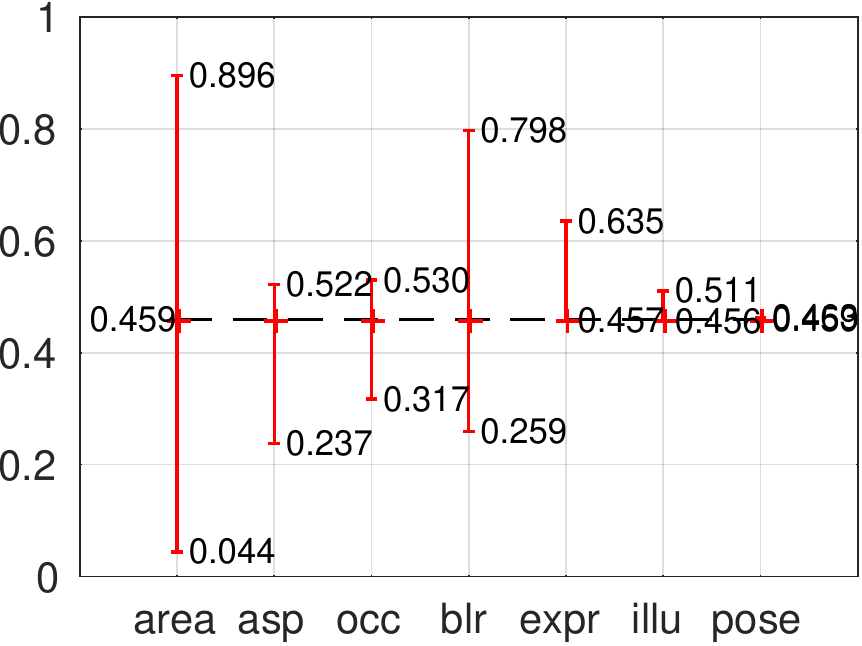}
  \caption{\label{fig:impact} Summary of sensitivity plot. We plot the maximum and minimum of $AP_N$ shown in Figure~\ref{fig:attribute}. Our detector is mostly affected by object scale (from 0.044 to 0.896) and blur (from 0.259 to 0.798).}
\end{figure}

\begin{figure*}
  \centering
  \includegraphics[width=.495\linewidth]{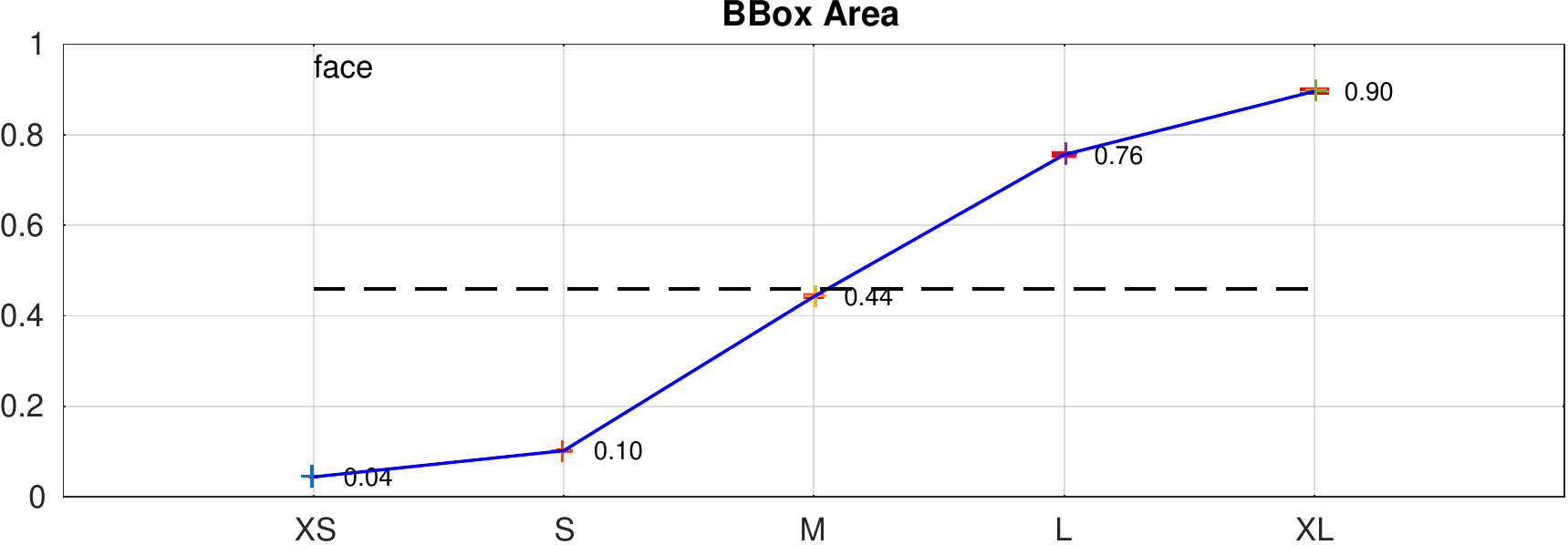}
  \includegraphics[width=.495\linewidth]{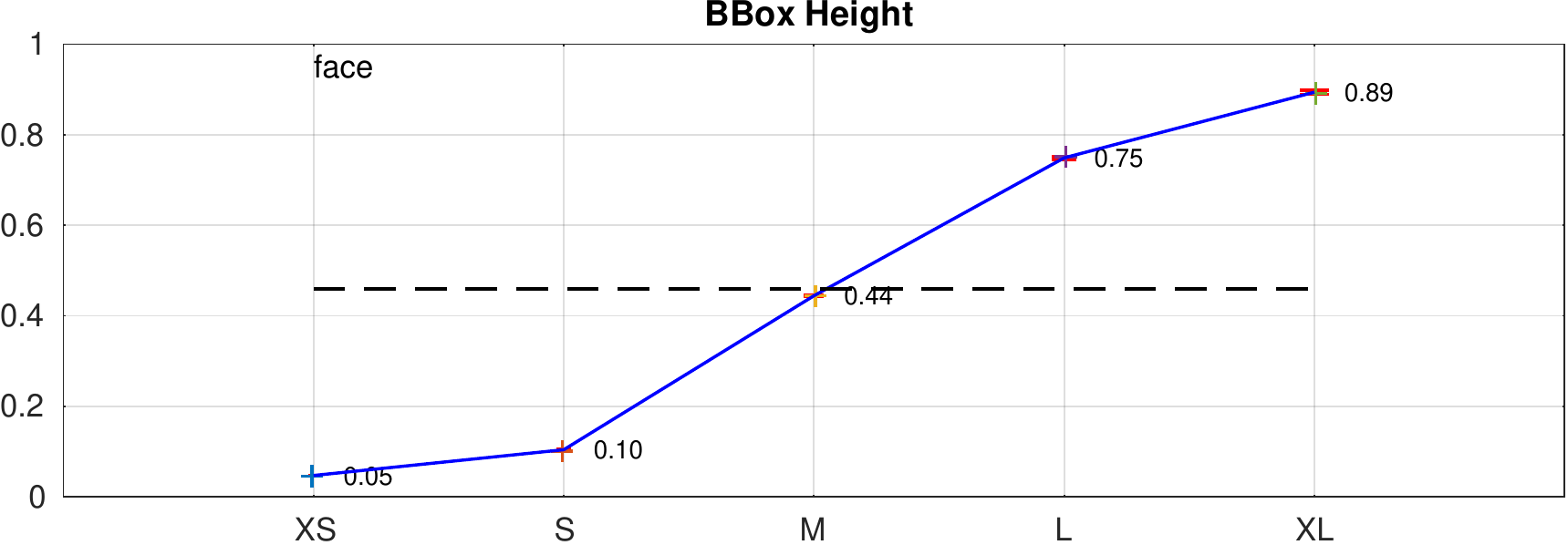}
  \includegraphics[width=.495\linewidth]{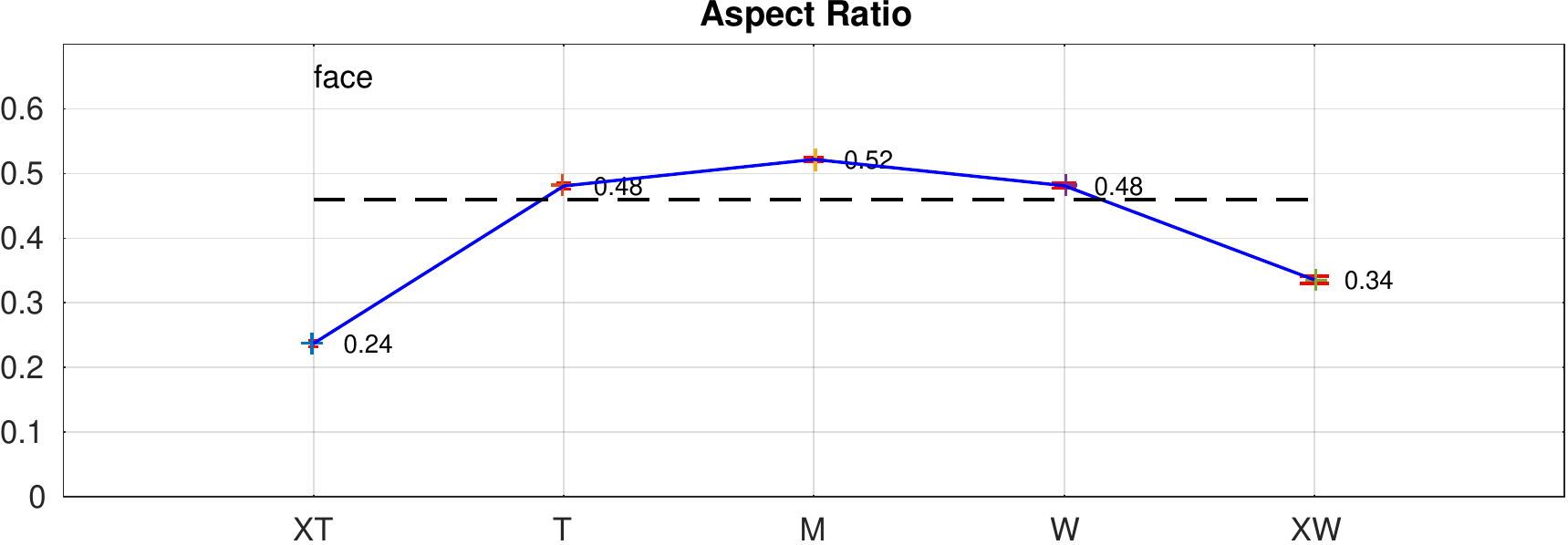}
  \includegraphics[width=.495\linewidth]{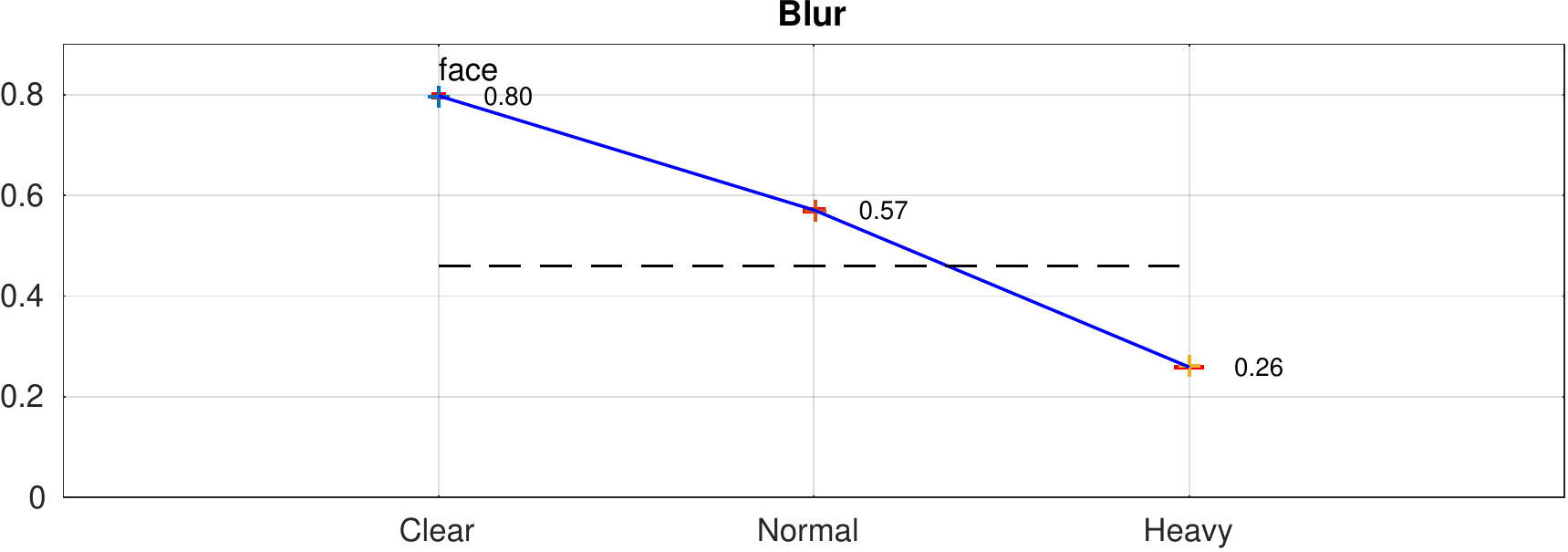}
  \includegraphics[width=.495\linewidth]{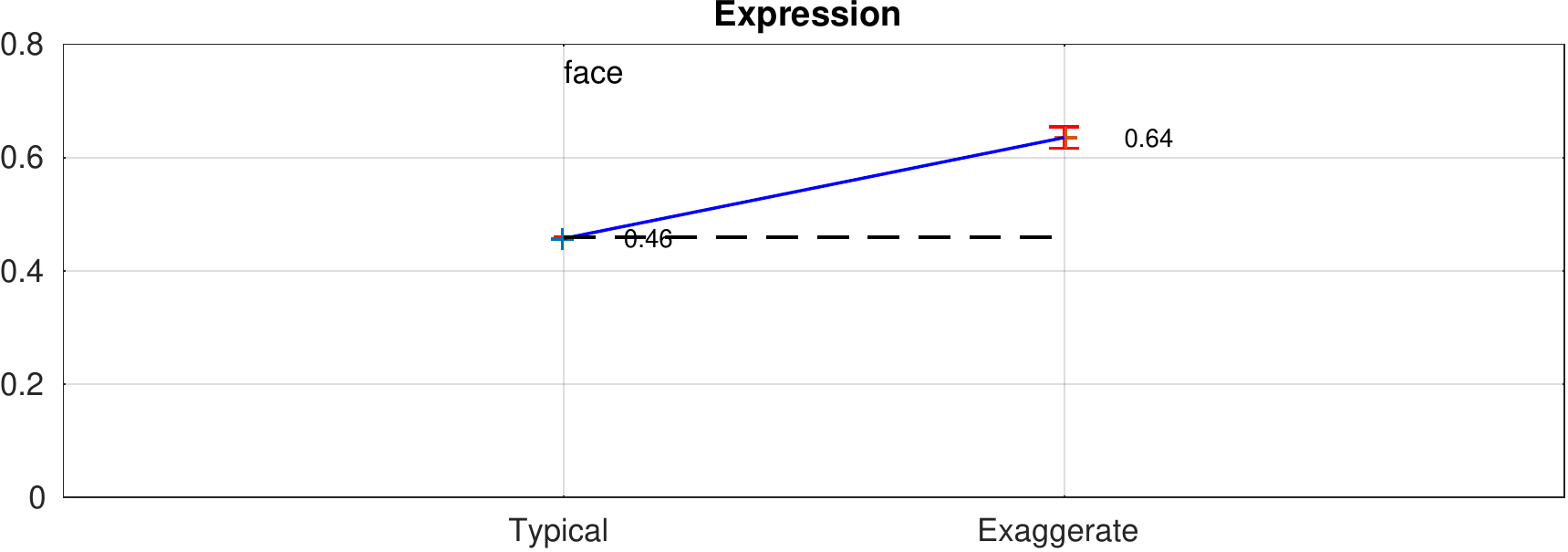}
  \includegraphics[width=.495\linewidth]{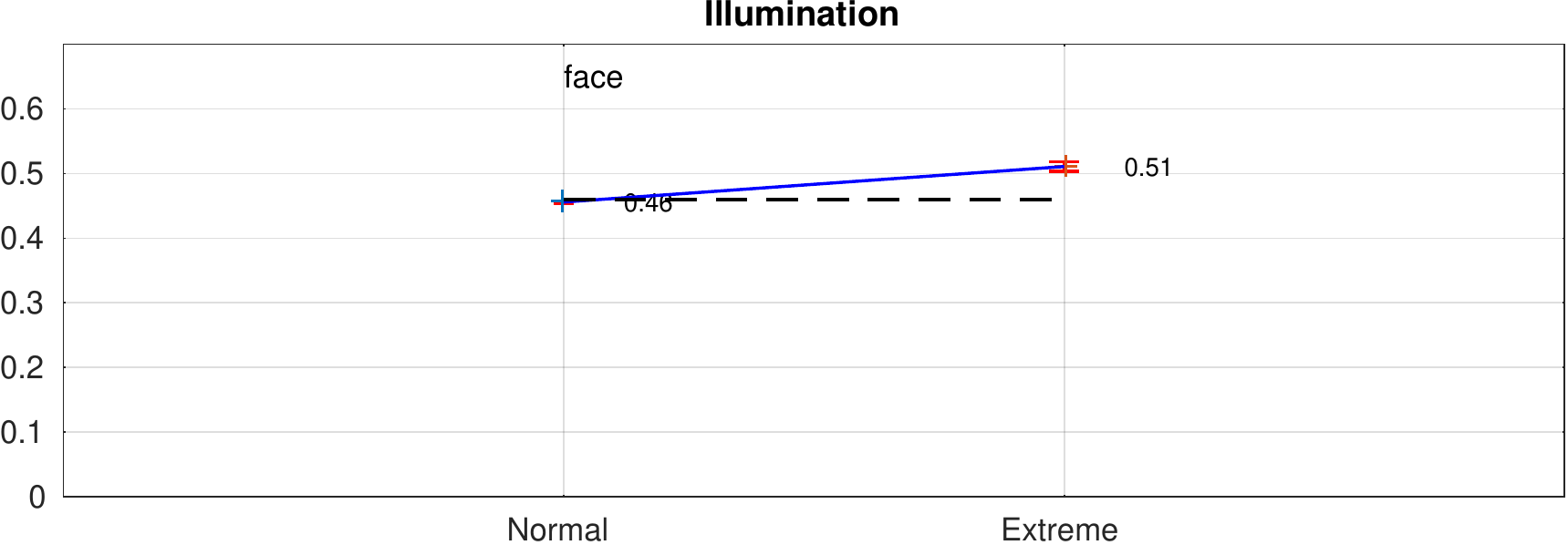}
  \includegraphics[width=.495\linewidth]{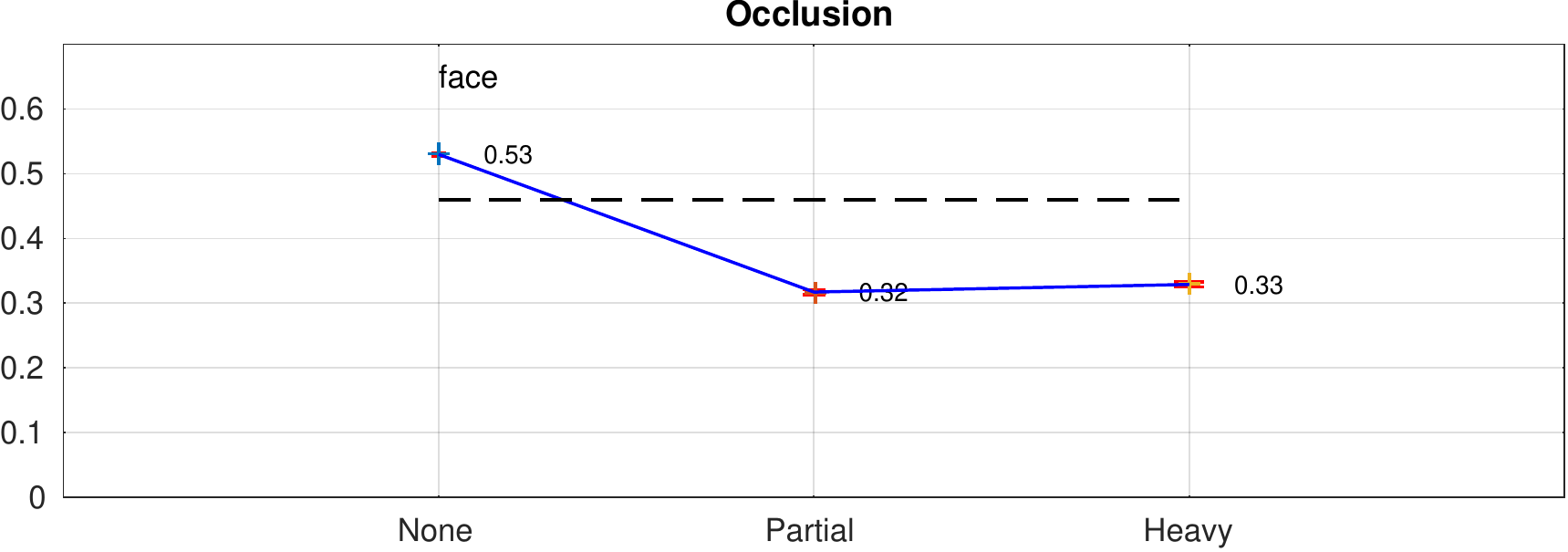}
  \includegraphics[width=.495\linewidth]{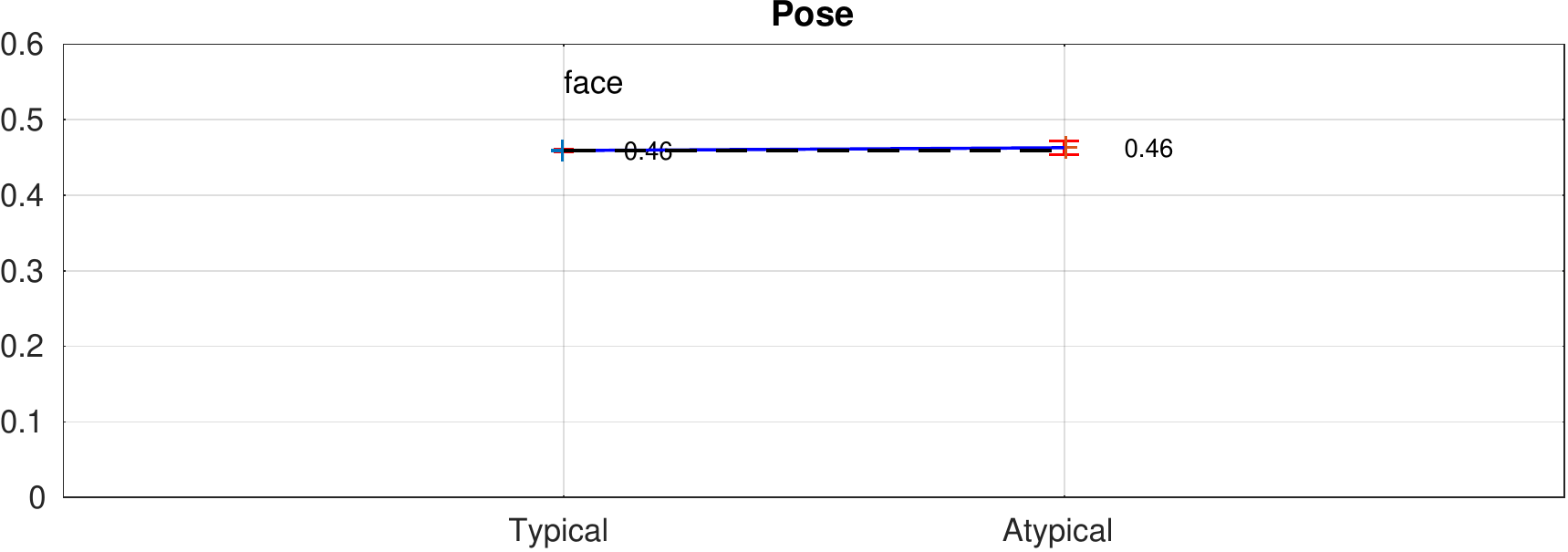}
  \caption{Sensitivity and impact of object characteristics. We show normalized AP\cite{hoiem2012diagnosing} for each characteristics. Please refer to \cite{hoiem2012diagnosing} for definition of ``BBox Area'', ``BBox Height'', and ``Aspect Ratio'' and also refer to \cite{yang2016wider} for the definition of per-face attributes ``Blur'', ``Expression'', ``Illumination'', ``Occlusion'', and ``Pose''. Our detector performs under average in the case of extremely small scale, extremely skewed aspect ratio, heavy blur, and heavy occlusion. Surprisingly, exaggerated expression and extreme illumination correlate with better performance. Pose variation does not have noticeable affect. }
  \label{fig:attribute}
\end{figure*}

{\bf Quantitative analysis}
We plot the distribution of error modes among false positives in Fig.~\ref{fig:error} and the impact of object characteristics on detection performance in Fig.~\ref{fig:impact} and Fig.~\ref{fig:attribute}. 

\begin{figure*}
  \centering
  \includegraphics[width=.87\linewidth]{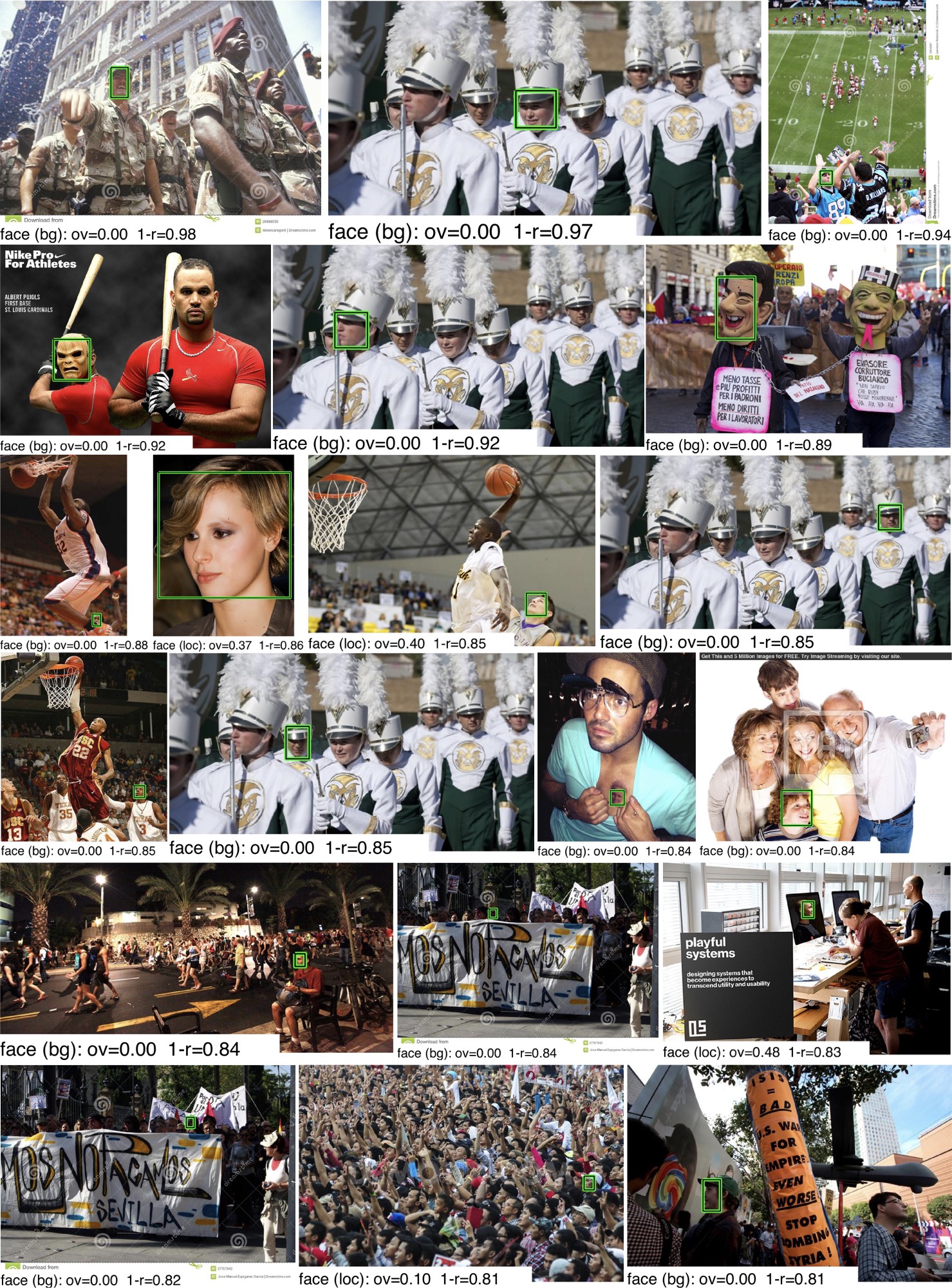}
  \caption{Top 20 scoring false positives on validation set. Error type is labeled at the left bottom of each image. ``face(bg)'' represents background confusion and ``face(loc)'' represents inaccurate localization. ``ov'' represents overlap with ground truth bounding boxes, ``1-r'' represents the percentage of detections whose confidence is below the current one's. Our detector seems to find faces that were not annotated (when prediction is on the face while ``ov'' equals to zero). }
  \label{fig:topfp}
\end{figure*}

{\bf Qualitative analysis}
We show top 20 scoring false positives in Fig.~\ref{fig:topfp}. 

\section{Experimental details}
\label{sec:experimental-details}
{\bf Multi-scale features} Inspired by the way \cite{shelhamer2016fully} trains ``FCN-8s at-once'', we scale the learning rate of predictor built on top of each layer by a fixed constant. Specifically, we use a scaling factor of 1 for res4, 0.1 for res3, and 0.01 for res2. One more difference between our model and \cite{shelhamer2016fully} is that: instead of predicting at original resolution, our model predicts at the resolution of res3 feature (downsampled by 8X comparing to input resolution). 

{\bf Input sampling} We first randomly re-scale the input image by 0.5X, 1X, or 2X. Then we randomly crop a 500x500 image region out of the re-scaled input. We pad with average RGB value (prior to average subtraction) when cropping outside image boundary. 

{\bf Border cases} Similar to \cite{ren2015faster}, we ignore gradients coming from heatmap locations whose detection windows cross the image boundary. The only difference is, we treat padded average pixels (as described in {\bf Input sampling}) as outside image boundary as well.


{\bf Online hard mining and balanced sampling} 
We apply hard mining on both positive and negative examples. Our implementation is simpler yet still effective comparing to \cite{shrivastava2016training}. We set a small threshold (0.03) on classification loss to filter out easy locations. Then we sample at most 128 locations for both positive and negative (respectively) from remaining ones whose losses are above the threshold. We compare training with and without hard mining on validation performance in Table~\ref{tab:hardmine}. 
\begin{table}
  \centering 
  \begin{tabular}{l|ccc}
    Method          & Easy  & Medium & Hard  \\
    \hline \hline
    w/  hard mining & {\bf 0.919} & {\bf 0.908}  & 0.823 \\
    w/o hard mining & 0.917 & 0.904  & {\bf 0.825} \\  
  \end{tabular}
  \caption{Comparison between training with and without hard mining. We show performance on WIDER FACE validation set. Both models are trained with balanced sampling and use ResNet-101 architecture. Results suggest hard mining has no noticeable affect the final performance. }
  \label{tab:hardmine}
\end{table}

{\bf Loss function} Our loss function is formulated in the same way as \cite{ren2015faster}. Note that we also use Huber loss as the loss function for bounding box regression.

{\bf Bounding box regression} Our bounding box regression is formulated as \cite{ren2015faster} and trained jointly with classification using stochastic gradient descent. We compare between testing with and without regression in terms of performance on WIDER FACE validation set. 
\begin{table}
  \centering 
  \begin{tabular}{l|ccc}
    Method         & Easy  & Medium & Hard  \\
    \hline \hline
    w/  regression & {\bf 0.919} & {\bf 0.908} & {\bf 0.823} \\
    w/o regression & 0.911 & 0.900  & 0.798 \\  
  \end{tabular}
  \caption{Comparison between testing with and without regression. We show performance on WIDER FACE validation set. Both models use ResNet-101 architecture. Results suggest that regression helps slightly more on detecting small faces (2.4\%).}
  \label{tab:reg}
\end{table}

{\bf Bounding ellipse regression} Our bounding ellipse regression is formulated as Eq.~\eqref{eq:er}. 
\begin{align}
  \label{eq:er}
  t^*_{x_c}  & = (x_c ^* - x_c ) / w  \\ 
  t^*_{y_c}  & = (y_c ^* - y_c ) / h  \\ 
  t^*_{r_a}  & = \log (r^*_a / (h/2)) \\
  t^*_{r_b}  & = \log (r^*_b / (w/2)) \\
  t^*_\theta & = \cot (\theta^*)      
\end{align}
where $x_c^*, y_c^*, r^*_a, r^*_b, \theta^*$ represent center x-,y-coordinate, ground truth half axes, and rotation angle of the ground truth ellipse. $x_c,y_c,h,w$ represent the center x-,y-coordinate, height, and width of our predicted bounding box. We learn the bounding ellipse linear regression offline, with the same feature used for training bounding box regression.

{\bf Other hyper-parameters} We use a fixed learning rate of $10^{-4}$, a weight decay of 0.0005, and a momentum of 0.9. We use a batch size of 20 images, and randomly crop one 500x500 region from the re-scaled version of each image. In general, we train models for 50 epochs and then select the best-performing epoch on validation set.

\clearpage

{
  \small
  \bibliographystyle{ieee}
  \bibliography{ref}
}

\end{document}